\documentclass[10pt,twocolumn,letterpaper]{xingyuanzhi}

\titlespacing*{\section}{0pt}{1.05\baselineskip plus 4pt minus 2pt}{0.35\baselineskip plus 2pt}
\titlespacing*{\subsection}{0pt}{0.75\baselineskip plus 3pt minus 1pt}{0.2\baselineskip plus 1pt}
\titlespacing*{\paragraph}{0pt}{0.75\baselineskip plus 2pt minus 1pt}{0.65em}
\setlist[itemize]{leftmargin=*,topsep=3pt plus 2pt minus 1pt,itemsep=2pt plus 1pt minus 1pt}
\setlist[enumerate]{leftmargin=*,topsep=3pt plus 2pt minus 1pt,itemsep=2pt plus 1pt minus 1pt}
\makeatletter
\let\xz@titleblock\mymaketitle
\renewcommand{\mymaketitle}{%
  \setbox0\vbox{\xz@titleblock}%
  \dimen0=\ht0 \advance\dimen0 by \dp0
  \count0=\dimen0 \count2=\baselineskip
  \divide\count0 by \count2 \advance\count0 by 1
  \dimen2=\count0\baselineskip \advance\dimen2 by -\dimen0
  \box0\vskip\dimen2}
\makeatother
\usepackage{xspace}
\usepackage{mathtools}
\usepackage{wrapfig}

\setcitestyle{numbers,square,comma}
\makeatletter
\def\NAT@sort{1}
\def\NAT@cmprs{1}
\makeatother

\renewcommand{\dblfloatpagefraction}{0.99}

\ifPDFTeX\else
\fi

\newcommand{\method}{\texttt{QuadArm-VLA}\xspace}

\newcommand{\headindent}{\leavevmode\hspace*{\parindent}}

\definecolor{rowhl}{HTML}{EFEFEF}

\definecolor{armc}{HTML}{B8610F}
\definecolor{legc}{HTML}{147565}
\newcommand{\arm}[1]{{\color{armc}#1}}
\newcommand{\leg}[1]{{\color{legc}#1}}

\usepackage{xstring}
\makeatletter
\definecolor{demoboxgray}{HTML}{6B7280}
\newlength{\demoboxtitleheight}
\newcommand{\demobox@titleformat}{%
  \begingroup
  \frenchspacing
  \footnotesize\ttfamily\bfseries
  \spaceskip=0.25em plus 0.05em minus 0.03em
}
\newcommand{\demobox@title}[1]{%
  \StrSubstitute{#1}{: }{:\,}[\demobox@title@tmp]%
  \demobox@title@tmp
}
\newtcolorbox{demobox}[2][]{
  enhanced,
  float*=p,
  width=\textwidth,
  colback=white,
  colframe=demoboxgray,
  arc=3pt,
  boxrule=0.5pt,
  fontupper=\small,
  before skip=11pt,
  after skip=11pt,
  top=4pt,
  bottom=4pt,
  left=8pt,
  right=8pt,
  boxsep=0pt,
  underlay={%
    \begin{scope}
      \clip[rounded corners=3pt] (frame.south west) rectangle (frame.north east);
      \fill[demoboxgray]
        (frame.north west) rectangle ([yshift=-\demoboxtitleheight]frame.north east);
    \end{scope}
  },
  overlay={%
    \ifx&#2&\else
      \node[anchor=west, text=white]
        at ([xshift=8pt, yshift=-0.5*\demoboxtitleheight]frame.north west)
        {\demobox@titleformat\demobox@title{#2}\endgroup};
    \fi
  },
  top=\dimexpr\demoboxtitleheight+3pt\relax,
  #1
}
\makeatother

\definecolor{phaseA}{HTML}{76A9D2}
\definecolor{phaseB}{HTML}{A78BC5}
\definecolor{phaseC}{HTML}{D99778}
\definecolor{phaseD}{HTML}{91AE89}
\definecolor{phaseE}{HTML}{C58D9F}
\newtcolorbox{phasebox}[1]{
  enhanced,
  colback=#1!40!white,
  colframe=#1,
  boxrule=0.5pt,
  arc=2pt,
  left=5pt,
  right=5pt,
  top=1pt,
  bottom=1pt,
  boxsep=0pt,
  before skip=1pt,
  after skip=1pt,
}
\newcommand{\demofigwidthab}{0.99\linewidth}
\newcommand{\demofigwidthcd}{0.99\linewidth}
\newcommand{\demofigwidth}{0.99\linewidth}
\newcommand{\phaseitem}[3]{\begin{phasebox}{#1}\textbf{#2}\enspace #3\end{phasebox}}

\projectname{QuadArm-VLA}
\reporttype{Technical Report}
\projectlogo{}

\title{\texorpdfstring{\method}{QuadArm-VLA}: Shaping Vision-Language-Action Systems for General Quadrupedal Whole-Body Loco-Manipulation}

\author[1,3,4\ast\ddagger]{Zhiming Xu}
\author[2,3,4\ast\ddagger]{Xu Li}
\author[3]{Yi Zou}
\author[3]{Lei Zhao}
\author[3,4\dagger]{Jiawei He}
\affiliation[1]{Carnegie Mellon University}
\affiliation[2]{Beijing University of Posts and Telecommunications}
\affiliation[3]{XYZ Embodied AI}
\affiliation[4]{Beijing Academy of Artificial Intelligence (BAAI)}
\contribution[\ast]{Equal contribution}
\contribution[\dagger]{Project lead}
\contribution[\ddagger]{Work done during an internship at XYZ Embodied AI and BAAI}

\abstract{
Autonomous quadrupedal loco-manipulation systems remain underexplored. Extending humanoid vision-language-action (VLA) pipelines is not straightforward, as this composite embodiment has no direct kinematic counterpart from which motion references can be retargeted. This raises two open questions: how an arm-body-coordinated whole-body controller should be derived, and how it should be bridged with a VLA policy using an appropriate action representation. Our key insight is that coordination discovery and execution can be separated. We first train a whole-body controller that discovers arm-body coordination and serves as the teleoperation interface, then re-express this coordination in a decoupled controller whose structured interface enables precise execution. We subsequently adapt a pretrained VLA to predict this interface, and the resulting system, \method, reaches 91\% average task progress across five mobile, contact-rich, and constrained manipulation tasks. Real-world demonstrations are available at \url{https://quadarm-vla.github.io/}. Our results provide practical guidance for acquiring coordinated motion and constructing VLA action interfaces for embodiments without a natural kinematic counterpart, and we will open-source our code to aid future research on relevant systems.
}

\hypersetup{pdfauthor={Zhiming Xu, Xu Li, Yi Zou, Lei Zhao, Jiawei He}}

\begin{document}
\maketitle

\begin{figure*}[t]
  \centering
  \includegraphics[width=\textwidth]{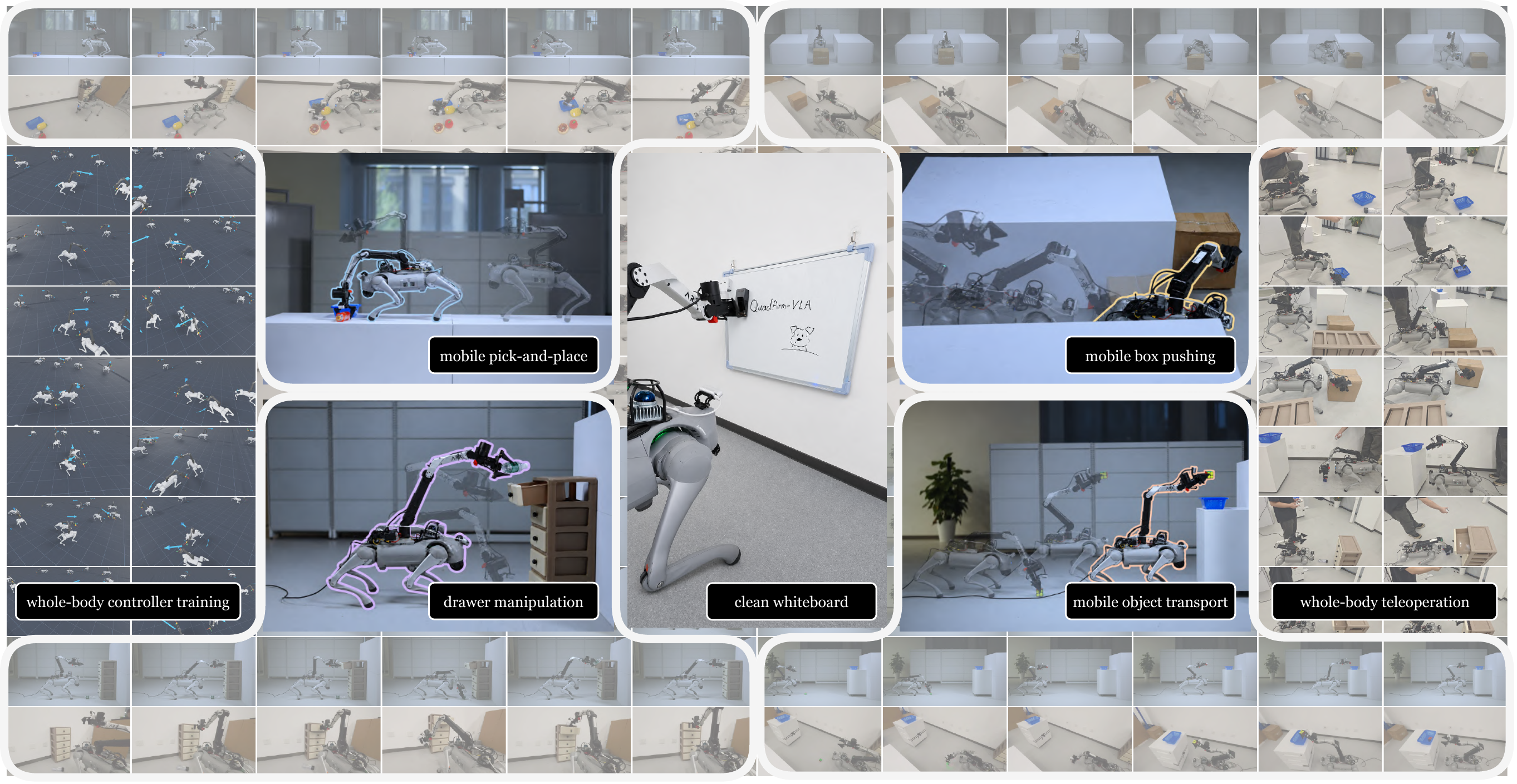}
  \caption{%
    \textbf{Overview of \method.}
    We obtain a whole-body motion generator in simulation (left), and collect whole-body demonstrations through VR teleoperation (right). A pretrained VLA is then adapted to predict the resulting structured controller interface.
    The center and surrounding panels show the resulting system on a Unitree Go2 with an ARX X5 arm executing language-conditioned tasks, including \textit{clean whiteboard}, \textit{mobile pick-and-place}, \textit{drawer manipulation}, \textit{mobile box pushing}, and \textit{mobile object transport}. Filmstrips along the borders show the temporal progression of each task. For additional real-world demonstrations of the modules described above, please visit our project website at \url{https://quadarm-vla.github.io/}.%
  }
  \label{fig:system-overview}
\end{figure*}

\section{Introduction}
\label{sec:intro}
\headindent Whole-body vision-language-action (VLA) systems, which pair a high-level VLA planner with a low-level whole-body controller, have enabled humanoids to perform a wide range of language-conditioned loco-manipulation tasks~\cite{wholebodyvla,psi-0,openhlm,sonic}. Human motion underlies both halves of these systems: retargeted motion-capture (MoCap) data trains the controllers to track reference motions~\cite{sonic,beyondmimic,openhlm}, and the same references form the action interface between the VLA and the controller.

We seek to build a comparable system for a quadruped equipped with a robotic arm. This composite embodiment combines legged mobility with an extended manipulation workspace, but it has no kinematic counterpart from which coordinated motion can be retargeted. Two questions thus remain underexplored: \emph{how to obtain references that capture arm-body coordination}, and \emph{how to expose this coordination through an action interface that a VLA can predict and execute precisely}. These questions distinguish our setting from humanoid whole-body VLA learning.

Our key insight is that coordination discovery and execution can be separated: the controller that discovers coordination need not be the one that executes VLA actions. To discover coordination, we train a whole-body reinforcement learning (RL) controller that maps end-effector poses and base velocities to all joints. Acting under the simulated dynamics, it produces feasible, coordinated arm-body motion, playing the role that MoCap plays for humanoids. Three training choices make it a stronger motion generator: simultaneous loco-manipulation training, separate value estimates for locomotion and manipulation, and reachable-pose augmentation of the command distribution~\cite{reachable-pose}. With them, the controller reduces end-effector position and orientation errors by 70\% and 81\% relative to RoboDuet~\cite{roboduet}. Its command space also makes teleoperation intuitive: the operator specifies a world-frame end-effector pose and a base velocity in VR, and the controller handles the rest.

The same end-effector interface, however, is ill-suited to VLA execution: a teleoperator watches the robot and corrects its centimeter-level residual errors and response delay, whereas a VLA executing action chunks cannot. To execute precisely, we re-express the discovered coordination in a decoupled controller, whose commands form a \emph{structured interface}: arm-joint targets, the arm-mount roll and pitch, and base velocity. A joint-position controller tracks the arm-joint targets with high precision and negligible latency, while a learned base policy tracks base velocity together with the roll and pitch, the two degrees of freedom (DoFs) through which the whole-body controller coordinates the arm and body. Because the base policy is trained on rollouts of the whole-body controller, it inherits both the coordinated postures and the transitions between them. Relative to the whole-body controller, it enlarges the effective reachable workspace by 22\% and reduces end-effector position error by 34\%.

On top of the structured interface, we fine-tune the pretrained VLA $\pi_{0.5}$~\cite{pi05} on teleoperated demonstrations to form \method (\Cref{fig:system-overview}). Because the base policy handles contact dynamics and leg stabilization, the VLA only predicts task-level arm-body commands. Adapting a VLA pretrained on wheeled dual-arm platforms to this embodiment still requires several ingredients: proprioception aligned with the structured interface and training-time action conditioning that keeps the command stream continuous during inference.

We instantiate \method on a Unitree Go2 with an ARX X5 arm and evaluate it on five real-world tasks covering mobile, contact-rich, and constrained manipulation. With the same VLA recipe, the structured interface achieves the highest progress on all five tasks, whereas the whole-body controller falls short on tasks that require a precise grasp or contact. Precision alone is not sufficient, however: the decoupled controller executes reliably only when trained on the whole-body controller's rollouts, demonstrating the necessity of both controllers. Using a single multi-task policy, \method reaches 91\% average progress, retains 68\% and 62\% under visual and object/layout changes, and, after mid-training on other tasks, learns a new task from fewer demonstrations. In summary, our contributions are fourfold:
\begin{itemize}
    \item \textbf{Whole-body RL for coordination discovery.} We formulate whole-body RL as a motion generator for composite embodiments without a kinematic counterpart, identify three training choices that improve it, and show that it also serves as an intuitive teleoperation interface.
    \item \textbf{A structured interface for precise execution.} We re-express the discovered coordination through the arm-mount roll and pitch in a decoupled controller, whose structured interface enlarges the effective reachable workspace, reduces tracking error, and enables precise VLA execution.
    \item \textbf{\method.} We introduce a whole-body VLA system built on this interface, and provide an empirical recipe for adapting a pretrained VLA to this embodiment.
    \item \textbf{Extensive real-world evaluation.} We validate our system on five real-world tasks, characterizing its visual robustness, generalizability and data efficiency.
\end{itemize}

\section{Related Works}
\paragraph{Whole-Body Controllers}%
Whole-body control coordinates all DoFs of a robot to extend its workspace and enable agile motion. Humanoid controllers typically exploit the structural correspondence between human and humanoid bodies, using retargeted human motion~\cite{deepmimic,h2o,omni-h2o,general-motion-retargeting,amass-human-mocap-data,lafan1-human-motion-data} as a coordination prior. RL then trains policies that track these reference motions~\cite{deepmimic,beyondmimic,sonic}, stabilize the lower body under human-like upper-body movements~\cite{mobile-television}, or coordinate a lower-body locomotion controller with commanded upper-body joint angles~\cite{wholebodyvla,leverb}. Without such a prior, controllers require task-specific reward design~\cite{homie,viral} and produce less diverse motions. Quadrupedal manipulators have no retargetable counterpart, so their whole-body coordination must be discovered rather than copied. Prior RL controllers for this embodiment take the form of a unified policy for the arm and base~\cite{deep-wbc,roboduet}, a pose-tracking controller over all DoFs~\cite{wheeled-pose-tracking,reachable-pose}, a multi-task trajectory tracker~\cite{umi-on-legs,mlm}, or a split between a model-based arm controller and a learned leg controller~\cite{relic,mpc-on-relic,visual-wbc,slim}. Our work bridges the unified and decoupled routes: we show that a unified interface is effective for teleoperation, whereas a decoupled interface derived from the learned coordination is better suited to executing VLA commands.

\paragraph{Whole-Body Teleoperation}%
Teleoperation supplies the demonstrations used to train high-level policies, and the controller's command interface determines what the operator must specify. Humanoid systems map operator body motion, captured by RGB cameras~\cite{h2o}, VR devices~\cite{omni-h2o,sonic,mobile-television,openhlm}, or exoskeletons~\cite{homie}, to reference motions for the whole-body controller, again relying on the human--humanoid correspondence. Quadrupedal manipulators lack this correspondence, so their teleoperation interface must be designed jointly with the controller, and existing designs each fall short: ReLIC~\cite{relic} requires whole-body inverse kinematics at deployment; body-frame end-effector commands~\cite{deep-wbc,roboduet} are sensitive to base perturbations and lack a fixed frame for forming action chunks; and handheld UMI grippers~\cite{umi}, as used in~\cite{umi-on-legs,mlm}, record end-effector poses but cannot easily encode base velocities. We instead let the operator specify a world-frame end-effector pose and a base velocity in VR, leaving coordination to the whole-body controller; the recorded poses are later converted into relative action chunks for VLA training.

\paragraph{Whole-Body VLA Systems}%
Automating whole-body controllers on real-world tasks requires a high-level vision-based planner that issues commands to the low-level controller. For quadrupedal manipulators, this planner has been learned from demonstrations~\cite{diffusion-policy,umi-on-legs}, built on systems with pretrained vision knowledge~\cite{quadwbg,odyssey,wildlma,helpful-doggybot}, or trained from scratch with RL~\cite{visual-wbc,slim}. However, these systems typically stack a planner and a skill library on top of the controller, and show limited generalization beyond their target tasks. Leveraging the language-following and generalization capabilities of VLA models~\cite{pi0,pi05,gr00tN1}, recent humanoid systems instead use a VLA as the planner, forming whole-body VLA systems~\cite{sonic,psi-0,openhlm,wholebodyvla,leverb}. The first is embodiment adaptation: whether a VLA or language-conditioned generalist policy pretrained on heterogeneous robot data can adapt to a target embodiment outside its training distribution~\cite{momanipvla,openhlm,x-vla}. The second is the planner--controller interface: LeVERB~\cite{leverb} and WholeBodyVLA~\cite{wholebodyvla} communicate through latent motions that encode coordination, other systems directly predict selected DoFs of the reference motion~\cite{sonic,openhlm,psi-0,pair-motion-with-text}, and recent works propose more abstract latent actions to accommodate heterogeneous action spaces~\cite{universal-action-token,cross-hand-latent}. Both questions remain largely unexplored for quadrupedal manipulators, whose systems stay task-specific owing to entangled low-level control and high-level planning~\cite{umi-on-legs} or task-specific design heuristics~\cite{visual-wbc,quadwbg,odyssey,fwbc-vla}. Our work addresses both questions for this embodiment: we adapt a VLA to a quadrupedal manipulator and give it an interface built for precise execution, unifying learned arm-body coordination with language-conditioned control.

\section{Method}
\begin{figure*}[t]
    \centering
    \includegraphics[width=\textwidth]{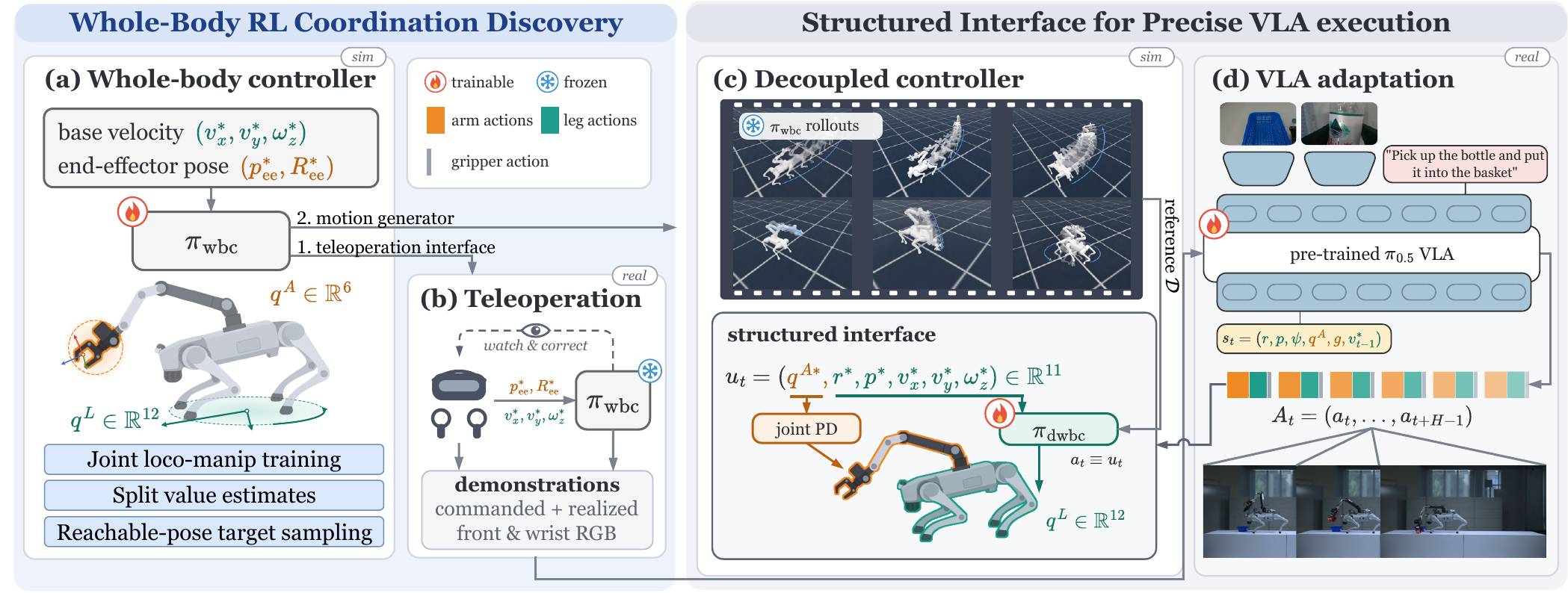}
    \caption{Overview of \method. \emph{Coordination discovery} (left). (a)~The whole-body controller $\pi_{\mathrm{wbc}}$ is trained in simulation to map base-velocity and end-effector pose commands to all 18 joints. It serves (b)~as the teleoperation interface, where the operator watches the robot and corrects its residual errors, and as a motion generator whose rollouts form the reference set $\mathcal{D}$. \emph{Precise execution} (right). (c)~The decoupled controller exposes the structured interface $u_t$: a joint PD controller tracks the arm targets $q^{A*}$ (orange), and $\pi_{\mathrm{dwbc}}$, trained to track $\mathcal{D}$, drives the legs (teal) to realize the arm-mount roll and pitch $(r^*,p^*)$ and the base velocity. (d)~$\pi_{0.5}$ is fine-tuned on the demonstrations to predict chunks $A_t$ of $u_t$ (plus a gripper command), which the decoupled controller executes on the real robot.}
    \label{fig:method}
\end{figure*}
\label{sec:method}

We consider a quadruped with 12 leg joints $\leg{q^L}$ and a 6-DoF arm $\arm{q^A}$ that performs language-conditioned whole-body loco-manipulation from a front and a wrist camera. \method separates the \emph{discovery} of arm-body coordination from its \emph{execution} (\Cref{fig:method}). A whole-body RL controller discovers the coordination and serves as the teleoperation interface (\Cref{sec:wbc,sec:teleop}). A decoupled controller then re-expresses this coordination through a structured interface that can be tracked precisely (\Cref{sec:dwbc}), and a pretrained VLA is shaped to predict this interface (\Cref{sec:vla}).

\subsection{Coordination Discovery}
\label{sec:wbc}
\headindent Collecting VLA demonstrations requires a controller that accepts end-effector poses and base velocities, and that recruits the legs into the arm's kinematic chain so that the demonstrations exhibit arm-body coordination. Since our embodiment has no kinematic counterpart from which coordination could be retargeted, we discover it with RL. The whole-body controller $\pi_{\mathrm{wbc}}$ is trained with PPO in simulation to track a command $c_t=(v^*,T^*_{\mathrm{ee}})$ with all 18 joints (\Cref{fig:method}a), where $v^*=(v^*_x,v^*_y,\omega^*_z)$ is the planar base velocity and yaw rate, and $T^*_{\mathrm{ee}}=(p^*_{\mathrm{ee}},R^*_{\mathrm{ee}})$ is a world-frame end-effector pose. Three training choices make it an effective motion generator:
\begin{itemize}[leftmargin=*,itemsep=1pt,topsep=2pt]
    \item \textbf{Joint loco-manipulation training.} Base-velocity and end-effector commands are sampled together, so the arm learns to track targets while the base is moving.
    \item \textbf{Split value estimates.} Separate critic heads fit locomotion and manipulation returns, disentangling credit assignment between base balance and arm tracking.
    \item \textbf{Reachable-pose target sampling.} End-effector targets are drawn from a precomputed set of collision-free reachable poses as in~\cite{reachable-pose}.
\end{itemize}

Because $\pi_{\mathrm{wbc}}$ hides whole-body coordination behind $c_t$, it plays two roles: it is the teleoperation interface (\Cref{sec:teleop}), and, since its rollouts obey the simulated dynamics, it generates feasible, coordinated motion for training the decoupled controller (\Cref{sec:dwbc}).

\subsection{Teleoperation Pipeline}
\label{sec:teleop}
\headindent A single operator drives all 18 DoFs through $\pi_{\mathrm{wbc}}$ with a VR headset and two handheld controllers (\Cref{fig:method}b). The joysticks set the base velocity, the trigger sets the gripper, and the right controller sets the end-effector pose through a clutch: pressing the grip anchors the current end-effector pose, the hand's subsequent motion moves the target at unit scale, and releasing the grip re-anchors it. The operator never specifies a joint, and corrects the residual errors of $\pi_{\mathrm{wbc}}$ by watching the robot.

Alongside the front and wrist images, we record both the operator's commands and the robot's realized state, namely the arm joints, arm-mount orientation and gripper position. Recording both lets the same demonstrations supervise either action interface: commanded end-effector poses for $\pi_{\mathrm{wbc}}$, or realized arm joints and arm-mount orientation, with the commanded base velocity, for the structured interface of \Cref{sec:dwbc}.

\subsection{Structured Action Interface}
\label{sec:dwbc}
\headindent A VLA executes its action chunks open loop, so its performance is bounded by how precisely and promptly the controller tracks them. $\pi_{\mathrm{wbc}}$ leaves residual end-effector errors and response delay, which a teleoperator corrects by watching but a VLA cannot. We therefore re-express its coordination through an interface that is easier to track.

Two observations shape this interface. First, the arm is easy to track and the base is not: a fixed-gain joint PD controller tracks arm joints precisely with negligible latency, whereas the base must handle contacts and counteract the forces and center-of-mass shifts induced by the arm. Second, $\pi_{\mathrm{wbc}}$ coordinates the arm and body mainly by rolling, pitching and yawing the base. Since yaw is already commanded through the yaw rate, we expose the roll and pitch of the arm mount as explicit commands.

The resulting decoupled controller (\Cref{fig:method}c) executes the structured interface
\begin{equation}
u_t=\big(\underbrace{\arm{q^{A*}}}_{\text{joint PD}},\;\underbrace{\leg{r^{*},\,p^{*},\,v^*_x,\,v^*_y,\,\omega^*_z}}_{\pi_{\mathrm{dwbc}}}\big)\in\mathbb{R}^{11},
\label{eq:decoupled-interface}
\end{equation}
where $r^*$ and $p^*$ are the desired arm-mount roll and pitch. A joint PD controller tracks the arm targets $\arm{q^{A*}}$, and a learned policy $\pi_{\mathrm{dwbc}}$ drives the legs to realize $(r^*,p^*)$ and the base velocity. Since $\pi_{\mathrm{dwbc}}$ observes the realized arm joints, it experiences the same coupled leg-arm dynamics as $\pi_{\mathrm{wbc}}$.

We train $\pi_{\mathrm{dwbc}}$ with PPO to track motion generated by $\pi_{\mathrm{wbc}}$, akin to motion tracking~\cite{beyondmimic,sonic}. We roll out the frozen $\pi_{\mathrm{wbc}}$ under sampled commands and record the realized arm joints, arm-mount roll and pitch, and base-velocity commands into a reference set $\mathcal{D}$, low-pass filtered to suppress oscillation. Each trajectory in $\mathcal{D}$ captures how $\pi_{\mathrm{wbc}}$ moves between coordinated postures, not only the postures themselves. A training episode starts from a state in $\mathcal{D}$ and replays its trajectory: the PD controller tracks its arm joints, while $\pi_{\mathrm{dwbc}}$ is rewarded for tracking its roll, pitch and base velocity (details in \Cref{app:reference-motions,app:controller-hyperparameters}).

This design leaves the arm to a precise joint controller, confines learning to the base, and transfers the coordination of $\pi_{\mathrm{wbc}}$ through $\mathcal{D}$. \Cref{sec:exp-coordination} shows that it enlarges the workspace and reduces end-effector error relative to $\pi_{\mathrm{wbc}}$.

\subsection{VLA Shaping}
\label{sec:vla}
\headindent \method fine-tunes the pretrained VLA $\pi_{0.5}$~\cite{pi05} to predict structured-interface actions from the two camera images, a proprioceptive state and a language instruction (\Cref{fig:method}d). Unlike the fixed or statically stable bases in its pretraining data, a quadrupedal manipulator poses three challenges: base posture reshapes the arm's reach and the camera view, the legged base must stay stable under VLA commands, and a stalled command stream can destabilize the moving robot. We address them with the input interface, output interface and training procedure, respectively.

\textbf{Input interface.} Besides the images, the VLA receives the proprioceptive state
\begin{equation}
s_t=\big(\leg{r_t,\,p_t,\,\psi_t},\,\arm{q^{A}_t},\,g_t,\,\leg{v^{*}_{t-1}}\big)\in\mathbb{R}^{13},
\label{eq:vla-proprioception}
\end{equation}
consisting of the arm-mount roll, pitch and episode-relative yaw, the arm joints, the gripper position, and the previous base-velocity command, a surrogate for the unmeasured base velocity. The arm-mount orientation tells the VLA how the base shapes the arm's reach and viewpoint. Leg joints are omitted, since gait is handled by $\pi_{\mathrm{dwbc}}$.

\textbf{Output interface.} Each action $a_t=(u_t,g^*)$ is the structured interface plus a gripper command, and the VLA predicts chunks $A_t=(a_t,\dots,a_{t+H-1})$. Because $\pi_{\mathrm{dwbc}}$ handles contacts and leg stabilization, the VLA only predicts task-level arm-body commands.

\textbf{Training.} We label each chunk with the realized arm joints, arm-mount roll and pitch, and gripper positions over the next $H$ steps, together with the commanded base velocities. Because a gap between chunks can destabilize the moving base, we fine-tune $\pi_{0.5}$ on the cleaned demonstrations with flow matching combined with training-time action conditioning~\cite{ttrtc}:
\begin{equation}
\begin{gathered}
\mathcal{L}=\mathbb{E}_{\tau,\epsilon,d}\Big[\big\lVert m_d\odot\big(v_\theta(X_\tau,\tau,o_t,\ell)-(\epsilon-A_t)\big)\big\rVert_2^2\Big],\\
X_\tau=m_d\odot\big(\tau\epsilon+(1-\tau)A_t\big)+(1-m_d)\odot A_t,
\end{gathered}
\label{eq:flow-matching}
\end{equation}
where $o_t$ is the observation, $\epsilon\sim\mathcal{N}(0,I)$, $\tau\in[0,1]$, and the mask $m_d\in\{0,1\}^H$ zeroes the first $d\sim\mathcal{U}(d_{\min},d_{\max})$ steps of the chunk. These prefix actions are given without noise, as if already committed, and are excluded from the loss. At deployment, the model thus generates the next chunk while the current one is still executing, so the robot always has a command to follow.

\section{Experiments}\label{sec:experiments}
The experiments are organized around five questions that follow the pipeline of \Cref{sec:method}.
\textbf{Q1:} Which design choices allow RL to effectively discover arm-body coordination for a quadrupedal manipulator?
\textbf{Q2:} Does re-expressing this coordination through the decoupled interface improve VLA execution, and what must the decoupled controller inherit from the whole-body controller?
\textbf{Q3:} What does adapting a pretrained VLA to this embodiment require?
\textbf{Q4:} How well does \method perform across tasks, and how robust is it to visual, object, and layout changes?
\textbf{Q5:} How many demonstrations does a new task require, and does mid-training on other tasks reduce this number?
\Cref{sec:exp-coordination} answers Q1--Q2, \Cref{sec:exp-vla-adaptation} answers Q3, and \Cref{sec:exp-capability} answers Q4--Q5.

\subsection{Experiment Setup}\label{subsec:experiment-setup}
\headindent\textbf{Controllers.} We model a Unitree Go2 equipped with an ARX X5 arm in IsaacLab~\cite{isaaclab}, train $\pi_{\mathrm{wbc}}$, roll it out to collect coordinated arm-body references, and train $\pi_{\mathrm{dwbc}}$ on these references. We evaluate tracking on the reachable~\cite{reachable-pose} and out-of-distribution (OOD) command set~\cite{roboduet}. We train and evaluate whole-body controllers on uneven terrain to support teleoperation across varied terrain conditions, whereas decoupled controllers use flat terrain with stronger randomization and disturbances, following common motion-tracking practice~\cite{beyondmimic,sonic}. We tune the reward configuration separately for each controller ablation and report its best observed evaluation score.

\textbf{Real-world system.} The physical platform carries one front and one wrist camera. Demonstrations are collected through VR teleoperation with $\pi_{\mathrm{wbc}}$, using LiDAR odometry from a Livox Mid-360S~\cite{fastlio2} to obtain world-frame end-effector poses.

\textbf{Tasks and metrics.} We evaluate five real-world tasks: \emph{mobile pick-and-place}, \emph{drawer manipulation}, \emph{mobile box pushing}, \emph{mobile object transport}, and \emph{clean whiteboard}, spanning mobile, contact-rich, and constrained manipulation. We decompose each task into ordered stages and report \emph{average task progress}~\cite{openhlm,pi0,pi05} over 10 trials. Training details are provided in \Cref{sec:app-wholebody-controller-training-details,sec:app-vla-training-details}, hardware specifications in \Cref{sec:app-teleoperation-data-collection}, and task stages and evaluation protocols in \Cref{sec:app-real-world-evaluation}.

\subsection{Coordination Discovery and Re-Expression}
\label{sec:exp-coordination}

\begin{table}[t]
    \centering
    \caption{Evaluation of whole-body controllers on uneven terrain (lower is better). LM: simultaneous loco-manipulation training; 2H: two-head value function; RS: reachable-set augmentation. Pos./Ori.: end-effector position and orientation errors; Lin./Ang.: base linear and angular velocity errors.}
    \label{tab:wbc-tracking}
    \footnotesize
    \setlength{\tabcolsep}{3.4pt}
    \begin{tabular}{lcccc}
        \toprule
        Method & Pos.\,(m) & Ori.\,(rad) & Lin.\,(m/s) & Ang.\,(rad/s) \\
        \midrule
        \multicolumn{5}{l}{\emph{Reachable command set}} \\
        \rowcolor{rowhl} $\pi_{\mathrm{wbc}}$ (ours) & \textbf{0.059} & \textbf{0.265} & \textbf{0.096} & 0.091 \\
        \quad w/o LM  & 0.191 & 0.610 & 0.154 & 0.095 \\
        \quad w/o 2H  & 0.155 & 0.580 & 0.117 & 0.100 \\
        \quad w/o RS  & 0.063 & 0.317 & \textbf{0.096} & \textbf{0.069} \\
        RoboDuet~\cite{roboduet} & 0.198 & 1.371 & 0.141 & 0.416 \\
        RoboDuet+RS   & 0.083 & 0.291 & 0.198 & 0.384 \\
        \midrule
        \multicolumn{5}{l}{\emph{OOD command set}} \\
        \rowcolor{rowhl} $\pi_{\mathrm{wbc}}$ (ours) & \textbf{0.185} & \textbf{0.744} & \textbf{0.104} & \textbf{0.096} \\
        RoboDuet      & 0.264 & 1.231 & 0.145 & 0.421 \\
        RoboDuet+RS   & 0.235 & 1.155 & 0.212 & 0.395 \\
        \bottomrule
    \end{tabular}
\end{table}

\begin{table}[t]
    \centering
    \caption{Evaluation of $\pi_{\mathrm{wbc}}$ and decoupled controllers on the reachable command set on flat terrain. ERW: effective reachable workspace (higher is better); remaining columns as in \Cref{tab:wbc-tracking} (lower is better).}
    \label{tab:dwbc-tracking}
    \footnotesize
    \setlength{\tabcolsep}{1.4pt}
    \resizebox{\linewidth}{!}{%
    \begin{tabular}{@{}lccccc@{}}
        \toprule
        Method & ERW\,(m$^3$) & Pos.\,(m) & Ori.\,(rad) & Lin.\,(m/s) & Ang.\,(rad/s) \\
        \midrule
        $\pi_{\mathrm{wbc}}$ (ours)  & 0.665 & 0.047 & 0.194 & 0.084 & 0.088 \\
        \rowcolor{rowhl} $\pi_{\mathrm{dwbc}}$ (ours) & \textbf{0.811} & \textbf{0.031} & 0.121 & 0.079 & 0.063 \\
        \quad w/o interpolation      & 0.800 & 0.033 & \textbf{0.120} & 0.073 & 0.053 \\
        \quad SLERP roll/pitch       & 0.769 & \textbf{0.031} & 0.121 & \textbf{0.069} & \textbf{0.050} \\
        \bottomrule
    \end{tabular}}
\end{table}

\headindent\textbf{Coordination discovery.} $\pi_{\mathrm{wbc}}$ outperforms RoboDuet on every metric under both command distributions (\Cref{tab:wbc-tracking}), reducing end-effector position and orientation errors on the reachable set by 70.4\% and 80.7\%. Reachable-set sampling improves RoboDuet's end-effector tracking at the cost of linear-velocity tracking, yet $\pi_{\mathrm{wbc}}$ remains better than RoboDuet+RS on all four metrics. The ablations show that removing simultaneous loco-manipulation training or the two-head value function more than doubles both end-effector errors, indicating that coordination benefits from experiencing arm and base motion jointly while keeping separate learning signals for the two objectives. Reachable-set augmentation further reduces end-effector errors at a small cost in yaw-rate tracking. In practice, prioritizing tracking accuracy too aggressively can lead to motions that overheat the motors, making such controllers unsuitable for teleoperation. We therefore include torque penalties and accept a loss in tracking accuracy; \Cref{tab:wbc-tracking} reports results for this regularized controller.

\textbf{Coordination re-expression.} The coordination of $\pi_{\mathrm{wbc}}$ survives its transfer to the decoupled interface (\Cref{tab:dwbc-tracking}). For decoupled controllers, end-effector targets are converted into arm-joint and arm-mount roll/pitch commands by an 8-DoF inverse kinematics (IK) solver. Effective reachable workspace (ERW) is calculated as the convex-hull volume of targets reached within $3\,\mathrm{cm}$ and $5^\circ$. Compared with $\pi_{\mathrm{wbc}}$, $\pi_{\mathrm{dwbc}}$ increases ERW by 22\% while reducing end-effector position and orientation errors by 34\% and 38\%, respectively. Because ERW counts only targets reached within the prescribed tolerances, this increase likely reflects more accurate tracking near the workspace boundary rather than expanded kinematic reach. The decoupled interface therefore preserves the evaluated reach of $\pi_{\mathrm{wbc}}$ while improving tracking precision.

\textbf{Reference construction.} $\pi_{\mathrm{dwbc}}$ is trained to follow the complete sequence of arm-joint angles and arm-mount roll and pitch that $\pi_{\mathrm{wbc}}$ traverses on its way to each target. We compare two variants that replace this transition. \emph{W/o interpolation} uses the terminal roll, pitch, and arm-joint angles directly as step targets, providing a more informed analogue of ReLIC-style target sampling~\cite{relic}. \emph{SLERP roll/pitch} instead interpolates the arm-mount orientation from its initial value to the target using spherical linear interpolation. In simulation, the transition matters little: all three decoupled controllers track end-effector poses almost equally well (3.1--3.3\,cm, 0.12\,rad). The two variants also track base velocity slightly more accurately, with only moderate reductions in workspace. These simulation metrics therefore provide little evidence for choosing among the reference constructions, motivating evaluation on the real robot.

\begin{figure}[t]
    \centering
    \includegraphics[width=\linewidth]{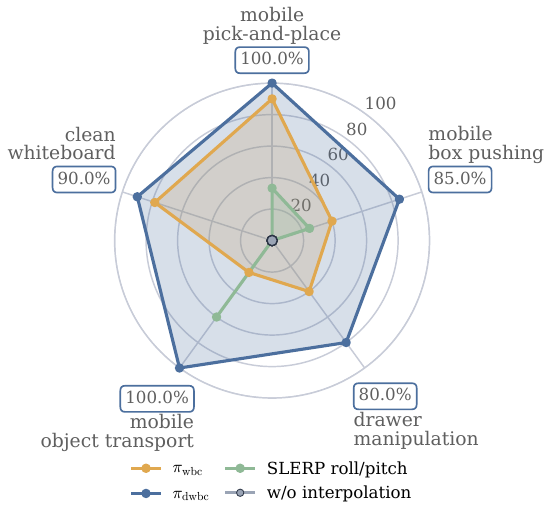}
    \caption{Real-world average task progress when the same VLA recipe is paired with each low-level controller. Labels give the progress of $\pi_{\mathrm{dwbc}}$.}
    \label{fig:action-interface-radar}
\end{figure}
\textbf{Downstream VLA execution.} Real-world VLA evaluation reveals substantial differences that are not captured by simulated tracking error (\Cref{fig:action-interface-radar}). With the same VLA recipe, $\pi_{\mathrm{dwbc}}$ attains the highest progress on every task. $\pi_{\mathrm{wbc}}$ remains competitive on mobile pick-and-place and clean whiteboard, which are more tolerant of small end-effector errors, but performs worse on drawer manipulation, mobile object transport, and mobile box pushing. In these tasks, $\pi_{\mathrm{wbc}}$ often misaligns the gripper with the object and pokes the box instead of pushing it, stalling the task. The two reference-construction variants perform substantially worse than their similar simulation errors suggest. The \emph{w/o interpolation} controller oscillates severely on the real robot and achieves zero progress on every task. Despite additional regularization tuning, the \emph{SLERP roll/pitch} controller produces aggressive body motions, including excessively low base postures, that destabilize the robot. The transitions produced by $\pi_{\mathrm{wbc}}$ avoid both failures because they are generated by a policy acting under the robot's dynamics. These results support separating the teleoperation interface from the execution interface, and show that the execution interface must inherit the transitions of $\pi_{\mathrm{wbc}}$ rather than merely using the tracking mode. Both failure patterns were consistently observed in all 10 independent trials.

\subsection{Embodiment Adaptation}
\label{sec:exp-vla-adaptation}

\begin{figure}[t]
    \centering
    \includegraphics[width=\linewidth]{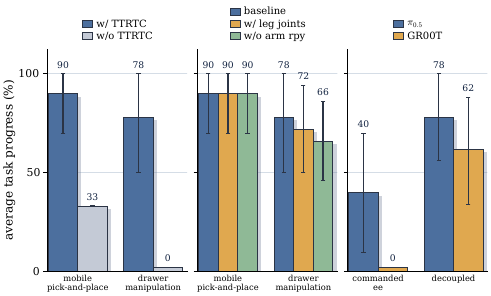}
    \caption{Average task progress for embodiment adaptation; error bars show 95\% confidence intervals. Left: training-time real-time chunking (TTRTC) ablation. Center: proprioceptive state design ablation. Right: comparison of pretrained VLA backbones for embodiment adaptation.}
    \label{fig:embodiment-adaptation}
\end{figure}

\begin{figure*}[t]
    \centering
    \includegraphics[width=\textwidth]{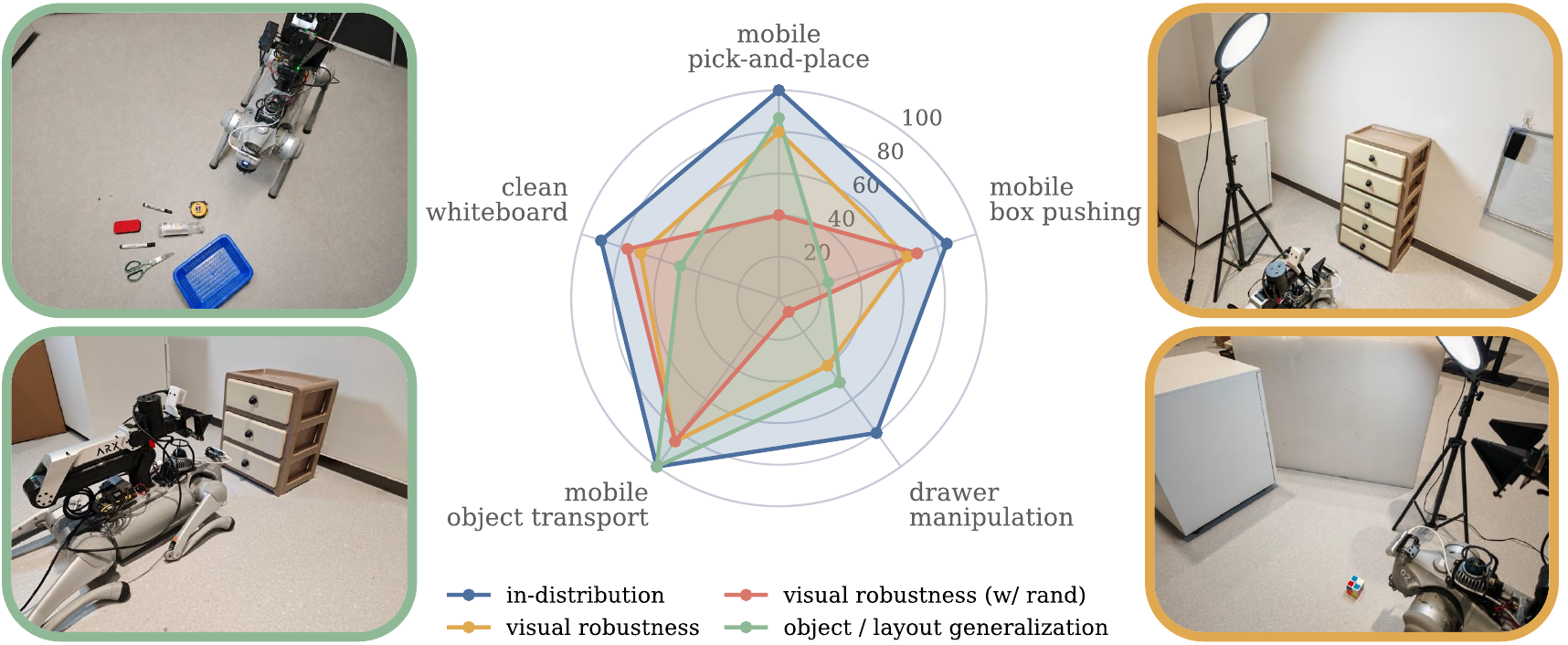}
    \caption{Visual robustness and object/layout generalization. Center: average task progress in distribution (blue), under strong lighting and background changes without (orange) or with (red; w/ rand) image randomization during training, and under object/layout changes (green). Left: representative layout changes, including distractor objects around the bottle in mobile pick-and-place (top) and a three-layer drawer replacing the five-layer drawer (bottom). Right: representative lighting changes.}
    \label{fig:robustness-generalization}
\end{figure*}

\headindent\textbf{Temporal continuity.} Continuous command streams are essential for a mobile base~\cite{psi-0}. Removing training-time real-time chunking (TTRTC)~\cite{ttrtc} lowers progress from 90\% to 33\% on mobile pick-and-place and from 78\% to 0\% on drawer manipulation (\Cref{fig:embodiment-adaptation}, left). Without it, inference gaps halt the moving base or force it to execute stale commands, and abrupt arm commands disturb the base during stationary manipulation.

\textbf{Proprioception.} Proprioception aligned with the decoupled interface works best, although the margins are small (\Cref{fig:embodiment-adaptation}, center). Our state (\Cref{eq:vla-proprioception}) reaches 90\% progress on mobile pick-and-place and 78\% on drawer manipulation. Neither variant changes mobile pick-and-place, but both lower drawer manipulation: adding the 12 leg joints reduces it to 72\%, likely because gait variables already handled by $\pi_{\mathrm{dwbc}}$ add high-frequency variation, and removing the arm-mount roll and pitch reduces it to 66\%, as the policy loses the feedback needed to adjust the arm-body workspace. These differences lie within the error bars, so we read them as supporting the state design rather than as decisive evidence.

\textbf{Backbone.} The decoupled interface improves both evaluated backbones on drawer manipulation (\Cref{fig:embodiment-adaptation}, right). Replacing end-effector commands with the decoupled interface increases task progress from 40\% to 78\% for $\pi_{0.5}$~\cite{pi05} and from 0\% to 62\% for GR00T N1.7~\cite{gr00tN1}. Nevertheless, GR00T underperforms $\pi_{0.5}$ with both interfaces. When predicting end-effector commands, GR00T often produces highly twisted poses that the arm cannot track reliably; when predicting joint targets, it produces larger and less stable action changes that disturb the compliant base. We therefore adopt $\pi_{0.5}$ as the backbone of \method.

\subsection{System Capability and Data Efficiency}
\label{sec:exp-capability}

\begin{figure}[t]
    \centering
    \includegraphics[width=\linewidth]{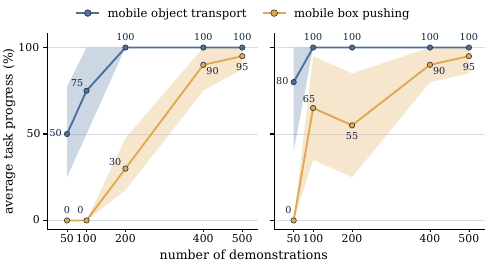}
    \caption{Data efficiency on mobile object transport and box pushing. Left: direct fine-tuning of $\pi_{0.5}$. Right: fine-tuning after mid-training on mobile pick-and-place and drawer manipulation. The $x$-axis shows target-task demonstrations; shading indicates 95\% confidence intervals.}
    \label{fig:data-efficiency}
\end{figure}

\headindent\textbf{Visual robustness.} All robustness and generalization experiments use a single multi-task policy trained on the five tasks. This policy achieves 91\% average in-distribution progress, which decreases to 68\% under strong lighting and background changes (\Cref{fig:robustness-generalization}, blue and orange). Four tasks retain 76--85\% of their in-distribution progress, whereas drawer manipulation drops from 80\% to 40\%. Randomizing image brightness, contrast, and saturation during training does not help and instead lowers average progress to 56\% (\Cref{fig:robustness-generalization}, red). We hypothesize that this augmentation degrades spatial cues needed for precise manipulation.

\textbf{Object and layout generalization.} We evaluate object changes by replacing the bottle with board eraser, sponge, or tissue (the language prompt accordingly), and by using a smaller box. For layout changes, we vary the drawer level or orientation, place the cube 20\,cm higher or lower, and raise, lower, or rotate the whiteboard. The combined results are shown in \Cref{fig:robustness-generalization} (green). Average progress is 62\%, with substantial variation across tasks: mobile object transport retains 100\% and mobile pick-and-place reaches 87\%, whereas drawer manipulation and clean whiteboard drop to 50\%, and mobile box pushing drops to 25\%. Failures concentrate on geometric changes: drawer manipulation drops to 25\% when the drawer level or orientation changes, while clean whiteboard reaches 67\% when the board is raised or rotated but only 17\% when it is lowered.

\textbf{Per-task data requirement.} Demonstration requirements vary substantially across tasks (\Cref{fig:data-efficiency}, left). When fine-tuned directly from $\pi_{0.5}$, mobile object transport reaches 100\% progress with 200 demonstrations. In contrast, mobile box pushing makes no progress with 50 or 100 demonstrations and requires 400 to reach 90\%. We hypothesize that this difference reflects the prevalence of pick-and-place data and the limited coverage of pushing behaviors in the $\pi_{0.5}$ pre-training mixture.

\textbf{Benefit of mid-training.} Mid-training on other tasks primarily improves adaptation in the low-data regime (\Cref{fig:data-efficiency}, right). We first mid-train $\pi_{0.5}$ on mobile pick-and-place (400 demonstrations) and drawer manipulation (500 demonstrations), then fine-tune it on each target task. For mobile object transport, mid-training increases progress from 50\% to 80\% with 50 target-task demonstrations and reaches 100\% with 100 demonstrations, compared with 200 under direct fine-tuning. For mobile box pushing, it achieves 65\% progress with 100 demonstrations, whereas direct fine-tuning makes no progress. The gains are not monotonic, possibly because source-task experience transfers to shared behaviors but provides limited support for task-specific pushing dynamics. Overall, mid-training reduces the target-task demonstrations required in the tested low-data settings without improving final performance.

\section{Conclusion}
\headindent We have presented \method, a whole-body VLA system for general quadrupedal loco-manipulation that addresses the fundamental challenge of operating composite embodiments without a natural kinematic counterpart. By training a whole-body controller in simulation and use it as the data collection engine, we collect whole-body teleoperation demonstrations. Our core insight is arm--body coordination discovery and interface for VLA execution can be decoupled, and with the resulting structured interface, we were able to construct a general whole-body VLA system enabling extensive loco-manipulation capabilities on a Unitree Go2 with an ARX X5 arm.

Despite the promising result and extensive real-world evaluation, our system has several limitations. Firstly, we rely on embodiment-specific data, and note leveraging robot-free data~\cite{wholebodyvla,gr00tN1} for quadrupedal manipulators remain a promising direction for future work. Secondly, we did not analyze the data efficiency of adapting our mid-trained model to other variants of quadrupedal manipulators. Nevertheless we open-source our system code to facilitate further research in the research community.

\bibliographystyle{ieeenat_fullname}
\bibliography{references}

\clearpage
\beginappendix
\counterwithin{equation}{section}
\counterwithin{figure}{section}
\counterwithin{table}{section}
\crefalias{section}{appendix}
\crefalias{subsection}{subappendix}
\crefalias{subsubsection}{subsubappendix}
\renewcommand{\dblfloatpagefraction}{0.8}

\twocolumn[{%
\section{Real-World Evaluation}\label{sec:app-real-world-evaluation}
\begin{demobox}[nofloat, before skip=4pt, after skip=\dbltextfloatsep]{Task 1: Mobile Pick-and-Place}

\textbf{Language Prompt:} Walk up to the bottle and basket, pick up the bottle from the ground, and put it into the basket.

\vspace{2pt}
\textbf{Task Description:} The robot walks to a suitable position near the bottle and basket, grasps the bottle from the ground, and places it into the basket. It must keep its balance throughout while coordinating the base, legs, and arm.

\vspace{2pt}
\textbf{Evaluation Rubric:} 3 stages in total:

\phaseitem{phaseA}{(a--e)}{The robot walks up to the correct position near the bottle and basket.}
\phaseitem{phaseB}{(f--g)}{The robot picks up the correct object (the bottle) from the ground.}
\phaseitem{phaseC}{(h)}{The bottle is placed into the basket.}

\vspace{2pt}
\textbf{Task Category:} mobile manipulation.

\vspace{2pt}
{\centering
\includegraphics[width=\demofigwidthab]{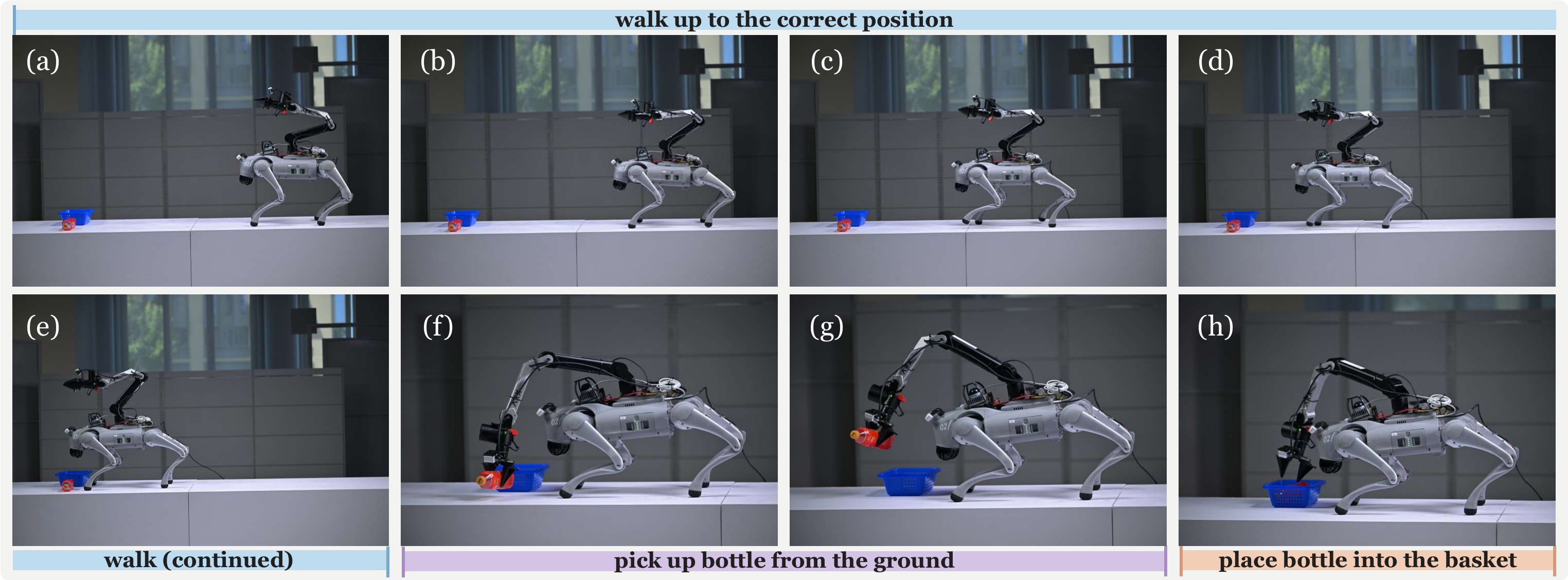}
\refstepcounter{figure}\label{fig:task1-description}
\par}
\end{demobox}
}]

\begin{demobox}{Task 2: Mobile Box Pushing}

\textbf{Language Prompt:} Clear the corridor by pushing the box behind the obstacle.

\vspace{2pt}
\textbf{Task Description:} The robot approaches the box, lowers the gripper to pushing height, and pushes the box sideways until it is behind the obstacle. It must keep the gripper in contact with the side of the box throughout the push, while the legs move the body and keep it stable against the reaction force of the box.

\vspace{2pt}
\textbf{Evaluation Rubric:} 4 stages in total:

\phaseitem{phaseA}{(a--b)}{The robot approaches and stops at a suitable position near the box.}
\phaseitem{phaseB}{(c--d)}{The robot positions its end effector at an appropriate height for pushing.}
\phaseitem{phaseC}{(e--f)}{The robot pushes the box away while maintaining stable contact.}
\phaseitem{phaseD}{(g--h)}{The box is pushed sideways behind the obstacle, clearing the corridor.}

\vspace{2pt}
\textbf{Task Category:} contact-rich manipulation.

\vspace{2pt}
{\centering
\includegraphics[width=\demofigwidthab]{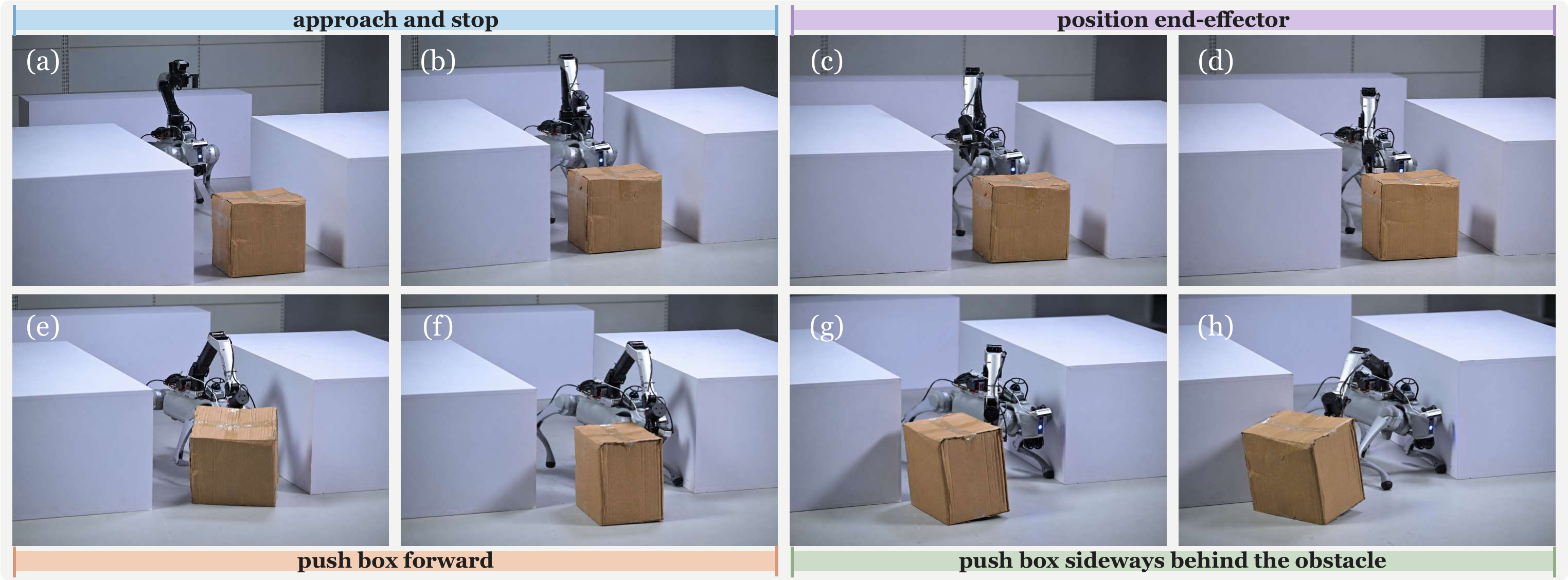}
\refstepcounter{figure}\label{fig:task2-description}
\par}

\end{demobox}

\begin{demobox}{Task 3: Drawer Manipulation}

\textbf{Language Prompt:} Open the top drawer, pick up the bottle from the ground, put it into the drawer, then close the drawer.

\vspace{2pt}
\textbf{Task Description:} The robot grasps the handle and pulls the top drawer open along its rails. It then picks up the bottle from the ground, places it inside the open drawer, and slides the drawer shut. The drawer rails constrain the gripper to move along a fixed direction, so the base, legs, and arm must move in coordination within this confined workspace.

\vspace{2pt}
\textbf{Evaluation Rubric:} 5 stages in total:

\phaseitem{phaseA}{(a)}{The robot grasps the circular handle of the drawer.}
\phaseitem{phaseB}{(b)}{The robot opens the top drawer.}
\phaseitem{phaseC}{(c--d)}{The robot grasps the bottle from the ground.}
\phaseitem{phaseD}{(e--f)}{The bottle is placed into the drawer.}
\phaseitem{phaseE}{(g--h)}{The drawer is closed.}

\vspace{2pt}
\textbf{Task Category:} constrained manipulation.

\vspace{2pt}
{\centering
\includegraphics[width=\demofigwidthcd]{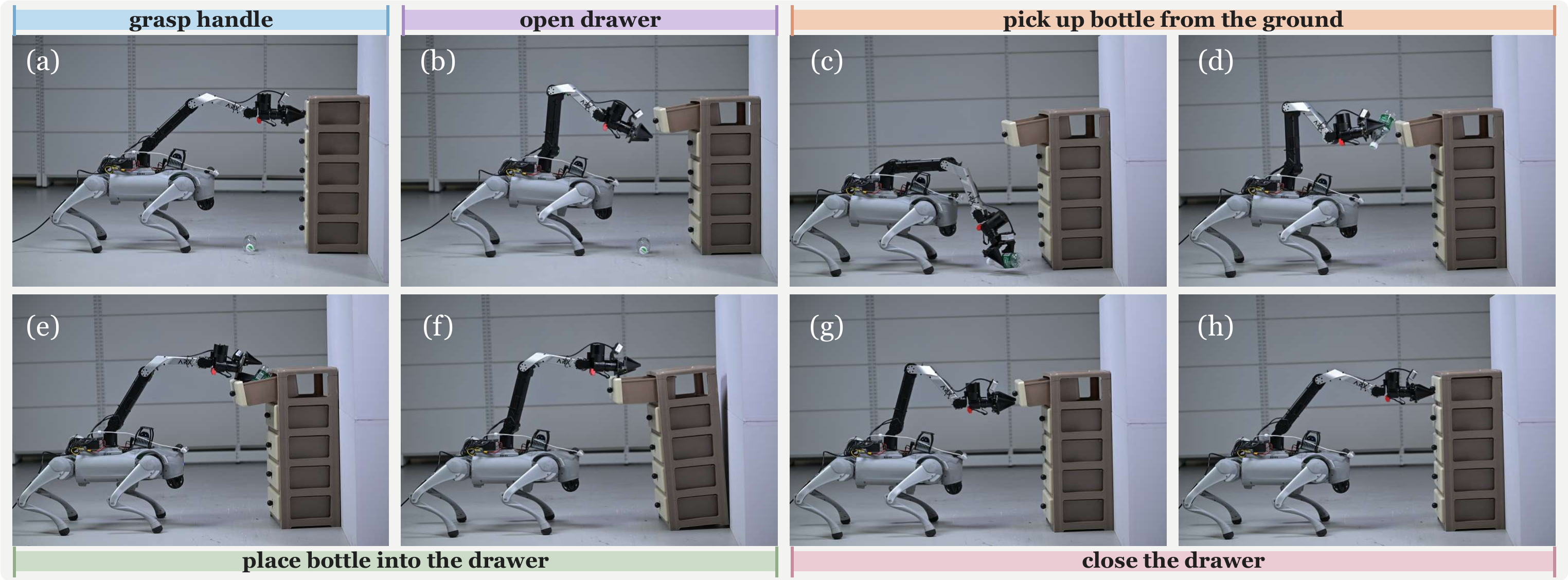}
\refstepcounter{figure}\label{fig:task3-description}
\par}

\end{demobox}

\begin{demobox}{Task 4: Mobile Object Transport}

\textbf{Language Prompt:} Pick up the cube from the ground, carry it forward, and put it into the basket.

\vspace{2pt}
\textbf{Task Description:} The robot grasps the cube from the ground and lifts it. It then walks forward while holding the cube, and finally positions the cube over the basket and releases it inside. During transport, the grasp must withstand the inertial disturbances caused by walking, so that the cube does not slip or fall.

\vspace{2pt}
\textbf{Evaluation Rubric:} 3 stages in total:

\phaseitem{phaseA}{(a--c)}{The robot grasps the cube from the ground.}
\phaseitem{phaseB}{(d--g)}{The robot carries the cube forward without dropping it.}
\phaseitem{phaseC}{(h)}{The cube is placed into the basket.}

\vspace{2pt}
\textbf{Task Category:} constrained mobile manipulation.

\vspace{2pt}
{\centering
\includegraphics[width=\demofigwidthcd]{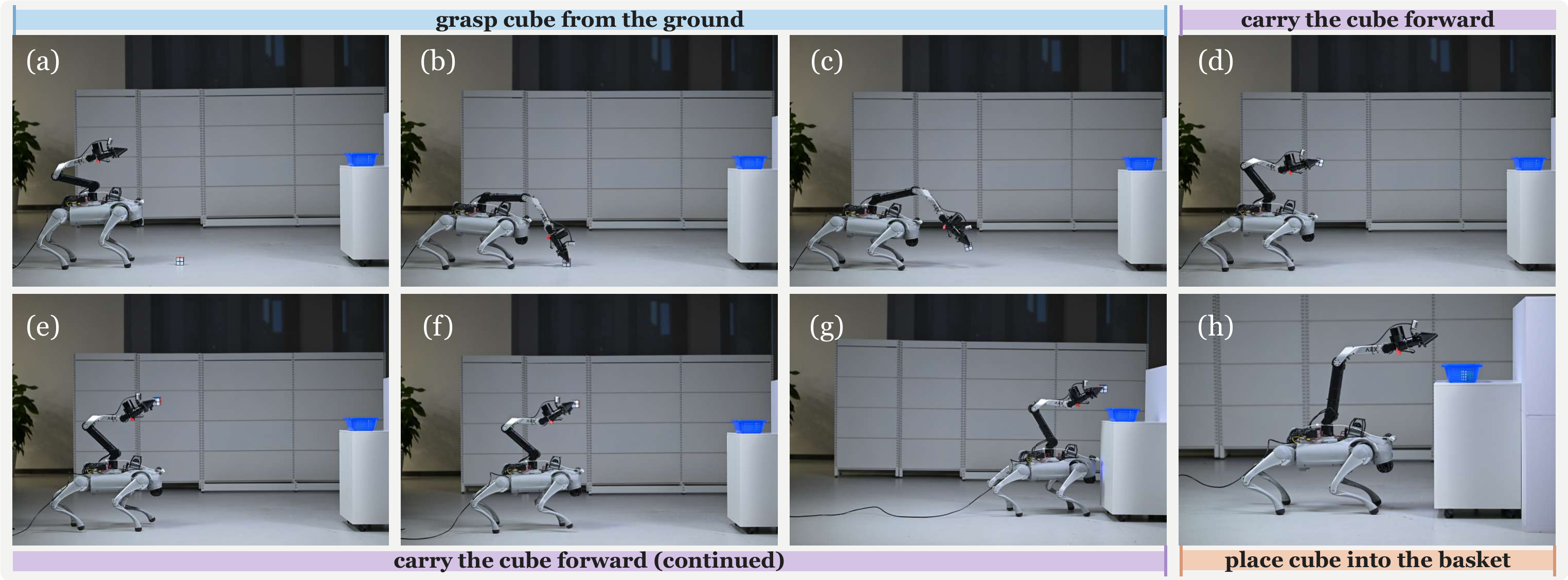}
\refstepcounter{figure}\label{fig:task4-description}
\par}

\end{demobox}

\begin{demobox}{Task 5: Clean Whiteboard}

\textbf{Language Prompt:} Pick up the wiper and erase the letters written on the whiteboard.

\vspace{2pt}
\textbf{Task Description:} The robot grasps the wiper from the table and lifts it to the whiteboard. It then wipes across the written letters while keeping the wiper in contact with the board, and finally returns the wiper to the table. Wiping requires continuous contact at the right height, without slipping off the board or dropping the wiper.

\vspace{2pt}
\textbf{Evaluation Rubric:} 6 stages in total, where \(x\) denotes the fraction of letters erased from the whiteboard:

\phaseitem{phaseA}{(a--b)}{The robot picks up the wiper.}
\phaseitem{phaseB}{(c--f)}{The robot attempts to wipe the letters, erasing \(0\% \leq x < 40\%\) of them.}
\phaseitem{phaseB}{(c--f)}{The robot erases \(40\% \leq x < 70\%\) of the letters.}
\phaseitem{phaseB}{(c--f)}{The robot erases \(70\% \leq x < 100\%\) of the letters.}
\phaseitem{phaseB}{(c--f)}{The robot erases all of the letters (\(x = 100\%\)).}
\phaseitem{phaseC}{(g--h)}{The robot places the wiper back on the table.}

\vspace{2pt}
\textbf{Task Category:} contact-rich manipulation.

\vspace{2pt}
{\centering
\includegraphics[width=\demofigwidth]{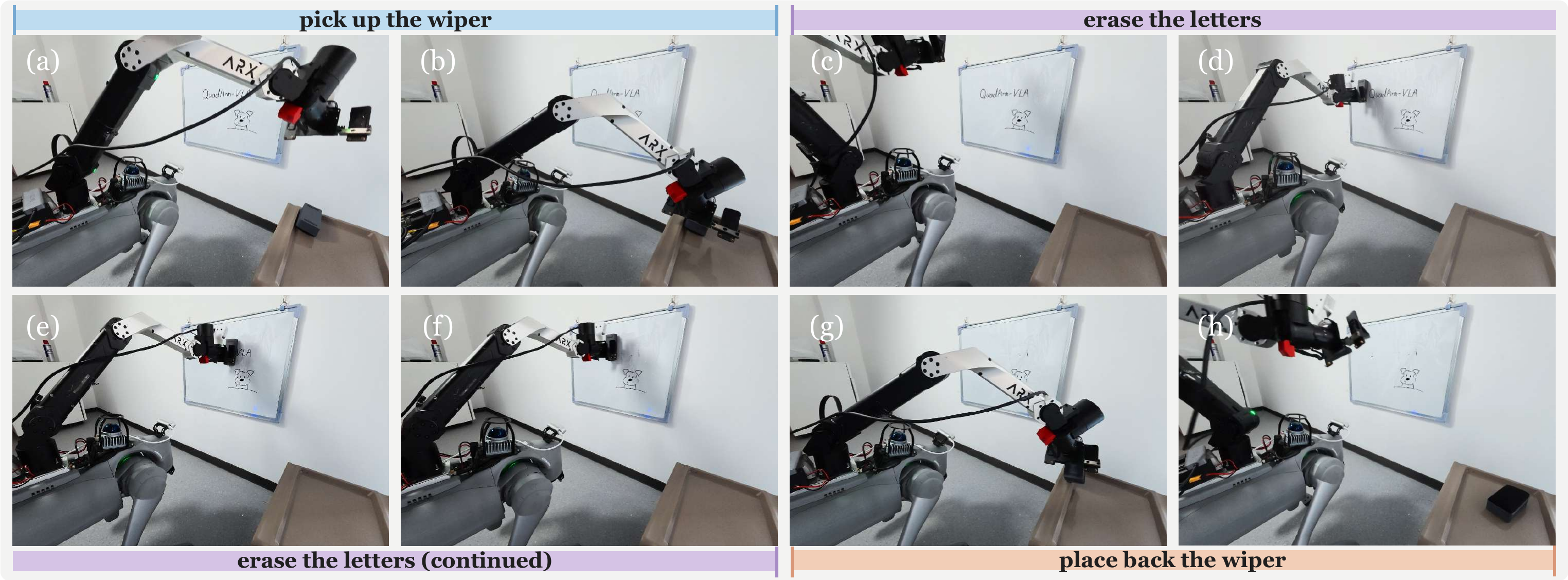}
\refstepcounter{figure}\label{fig:task5-description}
\par}

\end{demobox}

\subsection{Tasks and Evaluation Rubrics}\label{subsec:task-evaluation-protocol}
\headindent We evaluate \method on five real-world tasks, each described by one of the task cards in this section. A card gives the language prompt used at evaluation, the behavior the task requires, and a rubric of ordered stages. Each stage lists the filmstrip frames, labeled (a)--(h), in which it occurs, and is colored like the matching phase bar.

\subsection{Task Progress Metric}\label{subsec:app-task-completion-process}
\headindent We score each trial by its task progress. A trial earns one point for each stage of the task's rubric that it attains, so a trial that does not finish the task still receives partial credit. Because the tasks are sequential, a stage counts only if all earlier stages have been attained. In drawer manipulation, for example, grasping the bottle earns no point unless the drawer was opened first, and in clean whiteboard, picking up the wiper and putting it back without wiping earns only the first point. The progress of a trial is its number of points divided by the number of stages, which is 3 for mobile pick-and-place, 4 for mobile box pushing, 5 for drawer manipulation, 3 for mobile object transport, and 6 for clean whiteboard. We evaluate each task over 10 trials and report the average progress as a percentage, a metric commonly used to evaluate VLA policies~\cite{pi0,pi05,openhlm,gr00tN1}.

\section{Failure Case Analysis}\label{sec:app-failure-cases}
\begin{figure*}[t]
    \centering
    \includegraphics[width=\textwidth]{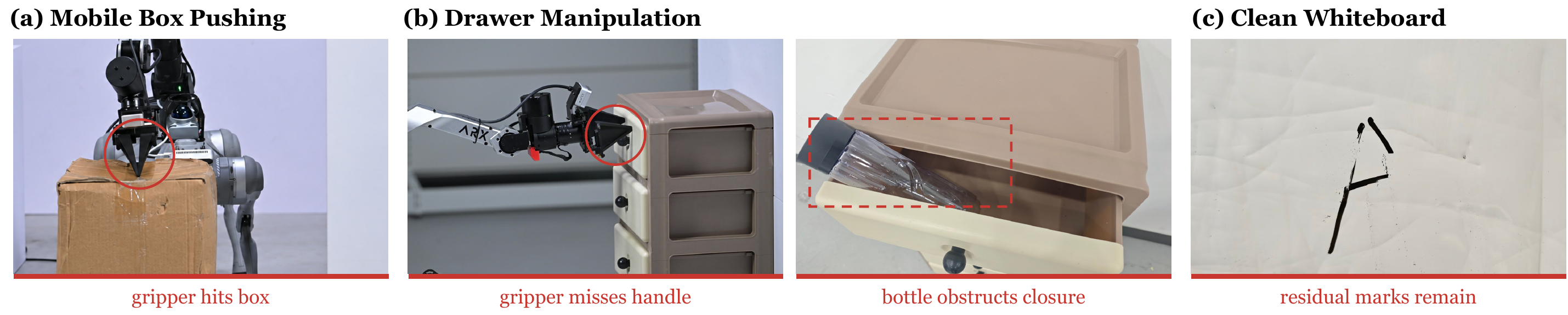}
    \caption{Representative failure cases of \method. (a)~Mobile box pushing: the gripper comes down on top of the box instead of pushing its side. (b)~Drawer manipulation: the gripper misses the handle (left), and the bottle is placed so that it blocks the drawer from closing (right). (c)~Clean whiteboard: fragments of the letters remain on the board after wiping.}
    \label{fig:failure-cases}
\end{figure*}
\headindent \method reaches 100\% average task progress on mobile pick-and-place and mobile object transport (\Cref{fig:action-interface-radar}). Both tasks center on picking and placing, which is likely well covered by the $\pi_{0.5}$ pre-training mixture. All failures therefore occur in the other three tasks, each with recurring failure modes.

\textbf{Mobile box pushing.} Pushing requires the gripper to meet the side of the box at a suitable height. In failed trials, the gripper instead comes down on top of the box (\Cref{fig:failure-cases}a), which stalls the push, and the task cannot proceed.

\textbf{Drawer manipulation.} This task fails in three ways. First, grasping the handle requires fine alignment, and a small offset makes the gripper miss it (\Cref{fig:failure-cases}b, left). Second, the robot sometimes places the bottle in a pose that blocks the drawer from closing (\Cref{fig:failure-cases}b, right). Third, the robot often reaches for the bottle before the drawer is open. This suggests that the policy does not condition the grasp on the drawer state and instead follows the most frequent behavior in the demonstrations, which points to adding a memory of task progress.

\textbf{Clean whiteboard.} The robot usually leaves small fragments of the letters on the board (\Cref{fig:failure-cases}c). We attribute this to perception: once the front and wrist images are resized to the VLA input resolution, these fragments span only a few pixels and are hard to detect.

\section{Additional Experiments}
\label{sec:app-additional-experiments}

\headindent We test whether an existing end-effector-centric whole-body controller, UMI-on-Legs~\cite{umi-on-legs}, can execute the end-effector commands of our VLA in place of our controllers. UMI-on-Legs tracks end-effector trajectories, and its original policy observes the scene only through a wrist camera, whose view moves with the end effector. Our tasks span a larger vertical workspace, from the ground to an elevated whiteboard, so our VLA also relies on a body-mounted front camera for scene context.

\textbf{Controller training.} We adapt UMI-on-Legs to our Go2--ARX X5 platform and train it on the stationary manipulation segments of our demonstrations, excluding locomotion, box pushing, and anomalous poses (\Cref{tab:uol-data,tab:uol-training}). The controller computes the end-effector pose from the base pose, which on our robot comes from LiDAR odometry with a latency of $90$--$120\,\mathrm{ms}$. A controller trained with low latency tracks more accurately (\Cref{tab:uol-tracking}), but on the real robot it overcorrects: it keeps correcting errors that the stale base pose still reports, and the resulting oscillations grow until the robot loses balance. We therefore train a latency-matched controller that receives base-pose updates at $25\,\mathrm{Hz}$, each delayed by $90$--$120\,\mathrm{ms}$, and use it for VLA integration.

\begin{table}[htbp]
    \centering
    \caption{Training data for the UMI-on-Legs controller, extracted from the stationary manipulation segments of each task.}
    \label{tab:uol-data}
    \footnotesize
    \setlength{\tabcolsep}{4pt}
    \begin{tabularx}{\linewidth}{@{}Xr@{}}
        \toprule
        Source task & Sequences \\
        \midrule
        Mobile pick-and-place & $1{,}122$ \\
        Mobile object transport & $702$ \\
        Drawer manipulation & $738$ \\
        Clean whiteboard & $307$ \\
        \midrule
        Total sequences & $2{,}869$ \\
        Total duration (hours) & $10.68$ \\
        \bottomrule
    \end{tabularx}
\end{table}

\begin{table}[htbp]
    \centering
    \caption{Training settings of the latency-matched UMI-on-Legs.}
    \label{tab:uol-training}
    \footnotesize
    \setlength{\tabcolsep}{4pt}
    \begin{tabularx}{\linewidth}{@{}l>{\raggedright\arraybackslash}X@{}}
        \toprule
        Parameter & Value \\
        \midrule
        Simulator / robot & Isaac Gym / Go2--ARX X5 \\
        Training algorithm & PPO, trained from scratch \\
        Parallel environments / seed & $4{,}096$ / $0$ \\
        Training iterations & $4{,}000$ \\
        Physics / reference timestep & $5\,\mathrm{ms}$ ($200\,\mathrm{Hz}$) \\
        Control frequency & $50\,\mathrm{Hz}$ \\
        Action-execution delay & $20\,\mathrm{ms}$ \\
        Base-pose update frequency & $25\,\mathrm{Hz}$ \\
        Base-pose observation delay & $90$--$120\,\mathrm{ms}$, sampled independently per pose packet \\
        Target observation & Current end-effector target only, no future frames \\
        Observation / action dimensions & $69$ / $18$ \\
        Action-rate penalty weight & $-0.1$ \\
        Terminal hold padding & None \\
        Evaluation checkpoint & Iteration $4{,}000$ \\
        \bottomrule
    \end{tabularx}
\end{table}

\begin{table}[htbp]
    \centering
    \caption{End-effector tracking errors of the two UMI-on-Legs controllers on fixed targets. Low latency: trained with low pose-feedback latency. LiDAR latency: the latency-matched controller of \Cref{tab:uol-training}.}
    \label{tab:uol-tracking}
    \footnotesize
    \setlength{\tabcolsep}{3pt}
    \begin{tabularx}{\linewidth}{@{}Xcc@{}}
        \toprule
        Controller & Position (cm) & Orientation ($^\circ$) \\
        \midrule
        Low latency & $1.01$ & $1.84$ \\
        LiDAR latency & $3.22$ & $4.66$ \\
        \bottomrule
    \end{tabularx}
\end{table}

\begin{figure}[htbp]
    \centering
    \includegraphics[width=\linewidth]{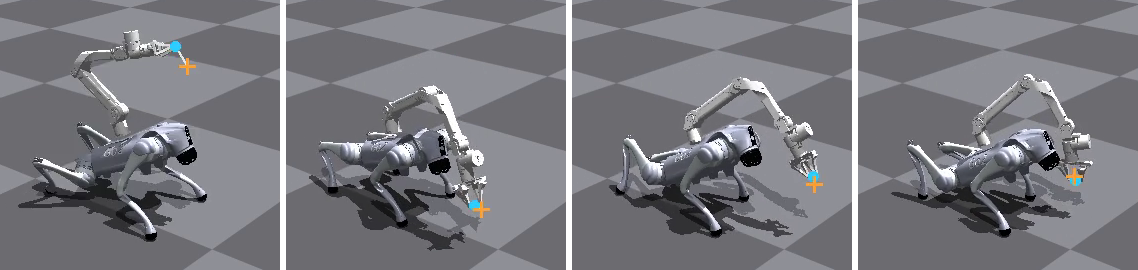}
    \caption{The latency-matched UMI-on-Legs controller tracking the end-effector trajectory of a mobile pick-and-place demonstration in simulation, from start (left) to end (right). The end effector (cyan) follows the target (orange), while the base stays nose-up.}
    \label{fig:uol-nose-up}
\end{figure}

\textbf{Nose-up posture.} The tracking itself works: the latency-matched controller follows a mobile pick-and-place trajectory to its end (\Cref{fig:uol-nose-up}). Throughout, however, it holds a pronounced nose-up posture, which we also observe on the real robot. The front camera then points away from the workspace and loses the scene context seen in the demonstrations, so the controller cannot be used with our VLA on the real robot: the VLA predicts erratic targets and completes none of the evaluated tasks. Stronger posture regularization might suppress this posture. However, the UMI-on-Legs implementation uses few regularization terms, and we found empirically that different tasks call for different regularization, whereas $\pi_{\mathrm{wbc}}$ uses the same terms for all end-effector targets. We therefore left its regularization untuned.

\textbf{Body posture as part of the interface.} An end-effector target does not determine the body posture, yet a body-mounted camera depends on it. Our demonstrations were collected with $\pi_{\mathrm{wbc}}$, so their front-camera views reflect its postures, which another end-effector controller need not reproduce. The structured interface of $\pi_{\mathrm{dwbc}}$ avoids this mismatch by commanding the arm-mount roll and pitch explicitly, so that the body posture, and with it the front-camera view, follows the demonstrations.

\section{Whole-Body Controller Training Details}\label{sec:app-wholebody-controller-training-details}
\headindent Both controllers are trained with PPO in simulation. $\pi_{\mathrm{wbc}}$ tracks end-effector targets drawn from the command pose sets of \Cref{app:command-pose-sets}, and $\pi_{\mathrm{dwbc}}$ tracks reference motions recorded from $\pi_{\mathrm{wbc}}$ (\Cref{app:reference-motions}). \Cref{app:controller-hyperparameters} lists the network, optimization, and reward settings of both controllers.

\subsection{Command Pose Set Generation}
\label{app:command-pose-sets}
\begin{figure}[ht]
    \centering
    \includegraphics[width=\linewidth]{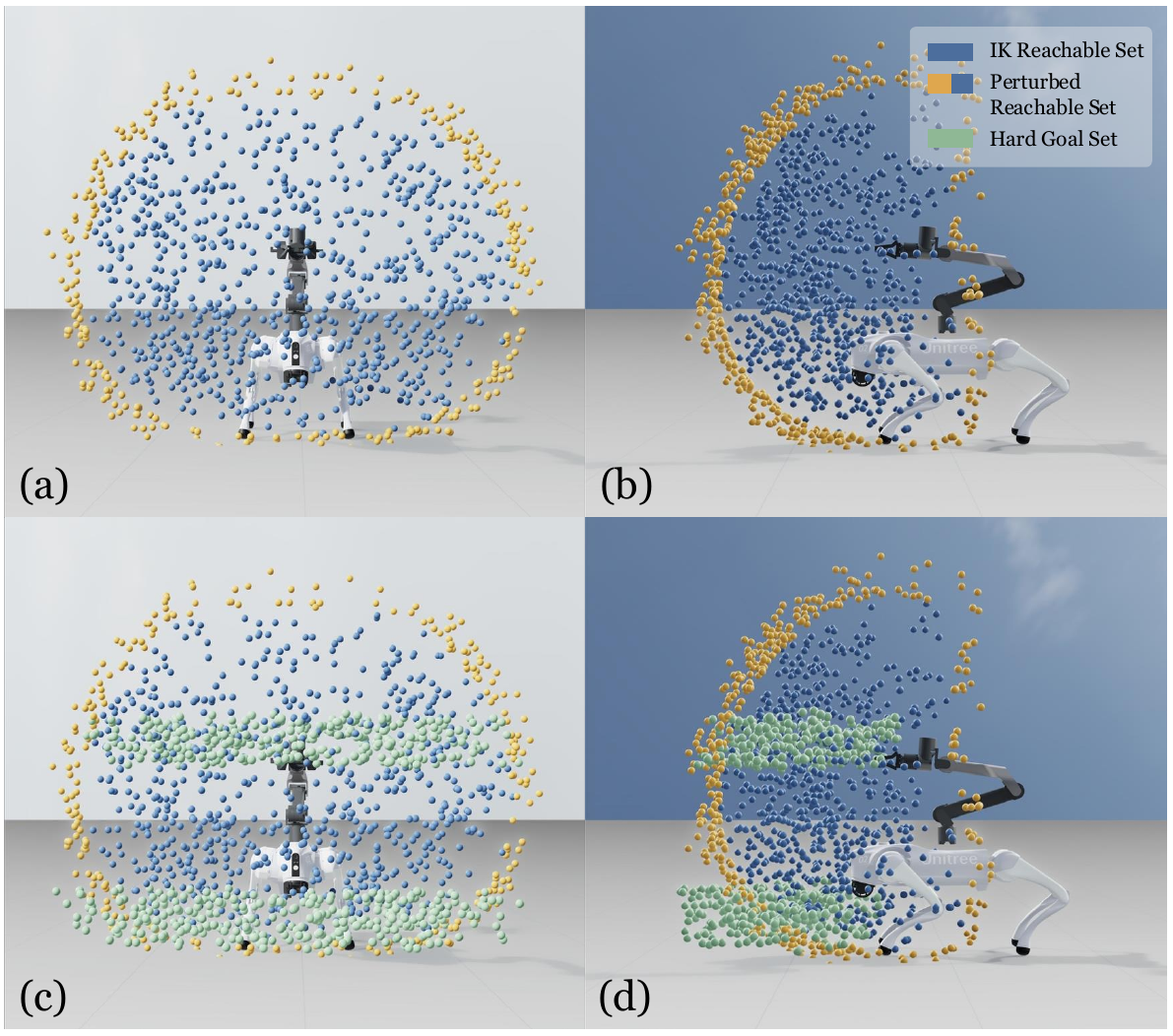}
    \caption{Command pose sets in front view (a, c) and side view (b, d). (a, b)~The IK reachable set (blue) and the additional targets of the perturbed reachable set (yellow). (c, d)~The hard goal set (green) added to the same views, forming a ground-pick band near the terrain and a high-reach band at shoulder height. Only a subset of the targets is shown.}
    \label{fig:command-pose-sets}
\end{figure}

\headindent Training $\pi_{\mathrm{wbc}}$ requires end-effector targets, but unlike humanoids, a quadrupedal manipulator has no retargeted human motion to provide them, and randomly sampled targets are often unreachable or in collision. We therefore construct two target sets (\Cref{fig:command-pose-sets}). Loco-manipulation episodes draw targets from a \emph{reachable set}, which improves end-effector tracking precision (\Cref{sec:exp-coordination}), and pure-manipulation episodes draw \emph{hard goals} that require the whole body. All targets are mapped to the world frame at every step through a heading frame, which follows the base position and yaw but stays at the nominal base height. The policy must therefore compensate for end-effector errors caused by base roll, pitch, and height changes.

\textbf{IK reachable set.} Following~\cite{reachable-pose}, we build a set of collision-free end-effector poses offline in simulation, with the base fixed at its nominal standing pose. We draw arm configurations from a scrambled Sobol sequence over the joint ranges in \Cref{tab:reachable-set}, which narrow joints 1, 2, 3, and 5 from their physical limits to exclude infeasible or unnatural postures. Each configuration is settled for two physics steps and rejected if any arm link registers a contact force above $1\,\mathrm{N}$, which removes collisions with the body and the ground. Forward kinematics then gives the end-effector pose $(\bm{p},\bm{R})$ in the base frame. Because uniform joint sampling concentrates poses in some regions of the workspace, a density filter rejects a candidate that lies close to at least two accepted poses. Sampling stops at $N=77{,}777$ poses, giving the set $\mathcal{S}=\{(\bm{p}_i,\bm{R}_i)\}_{i=1}^{N}$.

\textbf{Perturbed reachable set.} Since $\mathcal{S}$ is finite, we perturb its poses to cover the space between samples. When a loco-manipulation episode needs a new target, we draw $(\bm{p}_i,\bm{R}_i)$ uniformly from $\mathcal{S}$ and perturb it in the end-effector frame:
\begin{equation}
\bm{p}^{\star} = \bm{p}_i + \bm{R}_i\,\bm{\delta}_p, \qquad
\bm{R}^{\star} = \bm{R}_i \exp\!\big([\bm{\delta}_\theta]_\times\big),
\end{equation}
where $\bm{\delta}_p$ and $\bm{\delta}_\theta$ are drawn uniformly from 3D balls whose radii are given in \Cref{tab:reachable-set}. A new target is drawn every $3$--$6\,\mathrm{s}$.

\begin{table}[t]
\centering
\caption{Parameters of the reachable set.}
\label{tab:reachable-set}
\small
\setlength{\tabcolsep}{4pt}
\begin{tabularx}{\linewidth}{@{}>{\raggedright\arraybackslash}p{0.36\linewidth}>{\raggedright\arraybackslash}X@{}}
\toprule
Parameter & Value \\
\midrule
Joint ranges, joints 1--3 & $[-77,77]^\circ$, $[120,180]^\circ$, $[90,165]^\circ$ \\
Joint ranges, joints 4--6 & $[-90,90]^\circ$, $[-50,50]^\circ$, $[-90,90]^\circ$ \\
Sampler & Scrambled Sobol, $1{,}024$ per batch \\
Contact rejection & Force $>1\,\mathrm{N}$ on any arm link \\
Density filter & $\geq2$ accepted poses within $5\,\mathrm{cm}$ and $22^\circ$, after the first $300$ \\
Set size $N$ & $77{,}777$ \\
Perturbation radii & $3.7\,\mathrm{cm}$ (position), $7^\circ$ (orientation) \\
Resampling interval & $3$--$6\,\mathrm{s}$ \\
\bottomrule
\end{tabularx}
\end{table}

\textbf{Hard goal set.} Pure-manipulation episodes instead draw hard goals: far targets near the edge of the workspace that the arm alone cannot reach, so the policy must recruit the legs, which enlarges the workspace. Each goal belongs to one of two families with equal probability (\Cref{tab:hard-goal}): \emph{ground-pick} goals just above the terrain, reached by bending the front legs, and \emph{high-reach} goals at shoulder height, reached by bending the hind legs. A goal is written in spherical coordinates $(r,\phi,\psi)$ in the heading frame. We sample $r$, $\psi$, and a target height $h$ above the terrain, then solve for $\phi$ by bisection so that the goal sits at height $h$ above the terrain directly beneath it, using a ray cast against the terrain mesh. A goal is resampled if it falls inside the robot's enlarged bounding box, leaves the frontal cone, or violates its family's bounds, and a goal still invalid after $64$ rounds is replaced with a high-reach goal. The target orientation is sampled independently in the heading frame.

\begin{table}[t]
\centering
\caption{Hard goal sampling. $h$ is the target height above the terrain directly beneath the goal.}
\label{tab:hard-goal}
\small
\setlength{\tabcolsep}{4pt}
\begin{tabular}{@{}lccc@{}}
\toprule
Family & $r$ (m) & $h$ (m) & $\psi$ \\
\midrule
Ground-pick & $[0.48,0.92]$ & $[0.05,0.22]$ & $[-60^\circ,60^\circ]$ \\
High-reach  & $[0.48,0.92]$ & $[0.62,0.78]$ & $[-60^\circ,60^\circ]$ \\
\midrule
Orientation & \multicolumn{3}{l}{Roll, pitch, yaw $\sim\mathcal{U}[-60^\circ,60^\circ]$} \\
Frontal cone & \multicolumn{3}{l}{$x>0.12\,\mathrm{m}$, $|\psi|<60^\circ$} \\
Collision box & \multicolumn{3}{l}{Body bounding box enlarged by $0.1\,\mathrm{m}$} \\
Elevation solve & \multicolumn{3}{l}{Bisection, $10$ iterations} \\
\bottomrule
\end{tabular}
\end{table}

\subsection{Reference Motion Generation}\label{app:reference-motions}
\headindent $\pi_{\mathrm{dwbc}}$ learns to track reference motions, which we generate by running $\pi_{\mathrm{wbc}}$ in simulation and recording the motion it executes. The pipeline has three stages: rollout collection, reference extraction, and smoothing.

\textbf{Rollout collection.} We run $\pi_{\mathrm{wbc}}$ deterministically, using its mean action, at $50\,\mathrm{Hz}$ in $4{,}096$ parallel environments on flat terrain with $20\,\mathrm{s}$ episodes. We keep the startup randomization of friction, base mass, payload, and center of mass, but disable external pushes and end-effector force disturbances. Every $4$--$8\,\mathrm{s}$, each environment redraws its command and chooses locomotion or manipulation with equal probability, so a single episode can contain both. During locomotion, the policy receives a random base-velocity command, and its end-effector command holds the current end-effector pose in a frame that moves with the base. Short braking events at a rate of $0.1\,\mathrm{Hz}$ set the velocity command to zero for $0.3$--$0.8\,\mathrm{s}$. During manipulation, the velocity command is zero, and the end-effector target is a hard goal or a perturbed reachable-set pose (\Cref{app:command-pose-sets}) with equal probability. The target is fixed in the world frame at the base heading when it is sampled. At every step, we record the joint state, root state, end-effector pose, and commands.

\textbf{Reference extraction.} We split each environment's timeline into episodes and keep only those that reach the time limit, dropping episodes that end in a fall, an illegal contact, or excessive tilt. With $72{,}000$ steps per environment, this keeps $284{,}286$ episodes and drops $15{,}632$ failed ones ($5.2\%$). At each step, the reference contains the six arm joint positions, the arm-mount roll and pitch, and the base-velocity command. We store roll and pitch as a quaternion with the yaw removed, so the reference can be applied at any heading. We also store the full joint and root state, so that the tracking policy can start from any frame.

\textbf{Smoothing.} While $\pi_{\mathrm{wbc}}$ holds a target, its arm joints and arm-mount orientation oscillate slowly, and $\pi_{\mathrm{dwbc}}$ learns to reproduce these oscillations if trained on the raw trajectories. We therefore filter the six arm joint positions and the arm-mount roll and pitch, each channel independently, with a zero-phase Butterworth low-pass filter. Filtering proceeds in three steps, and the filter parameters are selected on a small pilot dataset of $243$ episodes collected with the same configuration. (1)~\textit{Phase segmentation.} A step is labeled as manipulation when $\|\bm{v}^{\mathrm{cmd}}\|<0.01$. Runs shorter than $1\,\mathrm{s}$ are merged into a neighboring run, so braking events are not mistaken for phase switches. (2)~\textit{Parameter selection.} Within each manipulation run, we split every channel into a \emph{reach} part and a \emph{hold} part, where the hold part begins once the channel stays within $0.03\,\mathrm{rad}$ over a trailing $1\,\mathrm{s}$ window. For each channel, we pool the hold parts, compute their power spectral density (PSD) with Welch's method, and set the cutoff to half the PSD's peak frequency. The filter order is the smallest that attenuates the peak frequency by at least $20\,\mathrm{dB}$, which is 2 for every channel. We raise a cutoff if needed so that the filter's step response settles within half of the 75th-percentile manipulation-run length, which was necessary only for pitch. (3)~\textit{Application.} Each manipulation run is filtered separately, never across a phase switch, and runs too short for the filter's transient are left unfiltered. During locomotion, we hold the arm joints at their last filtered values and leave roll and pitch unchanged. \Cref{tab:filter-cutoffs} lists the resulting cutoffs.

\begin{table}[h]
\centering
\small
\caption{Per-channel cutoff frequencies (Hz) of the second-order zero-phase Butterworth filter.}
\label{tab:filter-cutoffs}
\setlength{\tabcolsep}{5pt}
\begin{tabular}{cccccccc}
\toprule
Roll & Pitch & J1 & J2 & J3 & J4 & J5 & J6 \\
\midrule
0.17 & 0.11 & 0.27 & 0.46 & 0.42 & 0.75 & 0.40 & 0.40 \\
\bottomrule
\end{tabular}
\end{table}

\subsection{Training Hyperparameters}\label{app:controller-hyperparameters}
\headindent Both controllers are trained with PPO and the Adam optimizer, using the settings in \Cref{tab:controller-ppo-hyperparameters}. For $\pi_{\mathrm{wbc}}$, the actor and critic share a backbone with hidden widths $(512,256)$. The actor adds one 128-unit layer that outputs 18 action means, 12 for the legs and 6 for the arm, and the critic adds two 128-unit branches that output separate value estimates for locomotion and manipulation. For $\pi_{\mathrm{dwbc}}$, the actor and critic are separate MLPs with hidden widths $(512,256,128)$ that output 12 leg action means and one value estimate, respectively. All hidden layers use ELU activations, and both policies receive a five-step observation history. The Gaussian policies use learnable, state-independent log standard deviations, initialized to zero and clipped to $[-20,2]$, and the learning rate is annealed linearly from its initial to its final value. The two controllers share all optimization settings except the value-loss coefficient. The value loss of $\pi_{\mathrm{wbc}}$ is averaged over both heads, that is, over $2B$ values for a mini-batch of $B$ samples, so we double its coefficient to $2.0$ to keep the weight of each head equal to that of $\pi_{\mathrm{dwbc}}$.

\begin{table}[ht]
    \centering
    \small
    \caption{Network and PPO settings for the two controllers. Action scales convert policy outputs into joint-position offsets in radians.}
    \label{tab:controller-ppo-hyperparameters}
    \renewcommand{\arraystretch}{1.08}
    \setlength{\tabcolsep}{4pt}
    \begin{tabularx}{\linewidth}{@{}X>{\centering\arraybackslash}p{0.21\linewidth}>{\centering\arraybackslash}p{0.21\linewidth}@{}}
        \toprule
        Hyperparameter & $\pi_{\mathrm{wbc}}$ & $\pi_{\mathrm{dwbc}}$ \\
        \midrule
        Actor outputs / critic outputs & $18 / 2$ & $12 / 1$ \\
        Observation history & $5$ steps & $5$ steps \\
        Leg / arm action scale & $0.25 / 0.20$ & $0.25 / -$  \\
        Parallel environments & $4{,}096$ & $4{,}096$ \\
        Training steps per environment & $72{,}000$ & $72{,}000$ \\
        Rollout length per update & $24$ & $24$ \\
        PPO updates & $3{,}000$ & $3{,}000$ \\
        Epochs per update & $5$ & $5$ \\
        Mini-batches per epoch & $4$ & $4$ \\
        Samples per mini-batch & $24{,}576$ & $24{,}576$ \\
        Learning rate\newline\hspace*{1em}(initial $\rightarrow$ final) & $3\!\times\!10^{-4}$\newline$\rightarrow10^{-6}$ & $3\!\times\!10^{-4}$\newline$\rightarrow10^{-6}$ \\
        Discount factor $\gamma$ & $0.99$ & $0.99$ \\
        GAE parameter $\lambda$ & $0.95$ & $0.95$ \\
        Policy ratio clipping & $0.2$ & $0.2$ \\
        Value clipping & $0.2$ & $0.2$ \\
        Entropy loss coefficient & $0.01$ & $0.01$ \\
        Value loss coefficient & $2.0$ & $1.0$ \\
        Maximum gradient norm & $1.0$ & $1.0$ \\
        Reward scaling factor & $0.1$ & $0.1$ \\
        Time-limit bootstrapping & Enabled & Enabled \\
        Random seed & $42$ & $42$ \\
        \bottomrule
    \end{tabularx}
\end{table}

\begin{table}[ht]
    \centering
    \small
    \caption{Reward weights before the global reward scaling of \Cref{tab:controller-ppo-hyperparameters}. A dash marks an absent term. The gait contact-force and contact-velocity terms return non-positive values, so their positive weights penalize gait mismatch.}
    \label{tab:controller-reward-weights}
    \renewcommand{\arraystretch}{1.08}
    \begin{tabularx}{\linewidth}{@{}Xcc@{}}
        \toprule
        Reward term & $\pi_{\mathrm{wbc}}$ & $\pi_{\mathrm{dwbc}}$ \\
        \midrule
        Planar velocity tracking & $2.5$ & $1.5$ \\
        Yaw-rate tracking & $1.875$ & $0.75$ \\
        End-effector pose tracking & $2.0$ & $-$ \\
        Base roll/pitch tracking & $-$ & $1.5$ \\
        Feet air time & $2.5$ & $2.5$ \\
        Gait contact-force shaping & $2.0$ & $2.0$ \\
        Gait contact-velocity shaping & $2.0$ & $2.0$ \\
        Swing-foot clearance error & $-4.0$ & $-4.0$ \\
        Flat-base orientation penalty & $-2.5$ & $-2.5$ \\
        Leg posture deviation during locomotion & $-0.25$ & $-0.25$ \\
        Foot-contact reward during manipulation & $1.0$ & $1.0$ \\
        Standing leg posture deviation & $-0.2$ & $-0.2$ \\
        Base tilt tolerance during manipulation & $0.75$ & $0.75$ \\
        Base height error & $-5.0$ & $-5.0$ \\
        Vertical base velocity & $-2.0$ & $-2.0$ \\
        Base roll/pitch angular velocity & $-0.05$ & $-0.05$ \\
        Leg action rate & $-0.01$ & $-0.01$ \\
        Arm action rate & $-0.08$ & $-$ \\
        Leg joint torque & $-2\times10^{-4}$ & $-2\times10^{-4}$ \\
        Rear-hip joint torque (additional) & $-3\times10^{-4}$ & $-3\times10^{-4}$ \\
        Arm joint torque & $-5\times10^{-4}$ & $-$ \\
        Leg joint acceleration & $-2.5\times10^{-7}$ & $-2.5\times10^{-7}$ \\
        Arm joint acceleration & $-10^{-5}$ & $-$ \\
        Undesired body contacts & $-10.0$ & $-10.0$ \\
        Foot sliding & $-0.3$ & $-0.3$ \\
        Leg joint position limits & $-5.0$ & $-5.0$ \\
        Arm joint position limits & $-5.0$ & $-$ \\
        \bottomrule
    \end{tabularx}
\end{table}

\Cref{tab:controller-reward-weights} lists the reward weights. The two controllers share most gait and regularization terms. $\pi_{\mathrm{wbc}}$ additionally tracks the end-effector pose and regularizes the arm, whereas $\pi_{\mathrm{dwbc}}$ tracks the reference arm-mount roll and pitch while the arm follows the replayed joint targets. Terms marked as applying during locomotion or manipulation are active only when the command selects that mode. Both controllers use a $3\,\mathrm{Hz}$ trot reference with a stance fraction of $0.5$ and a target swing-foot clearance of $0.07\,\mathrm{m}$.

\section{Teleoperation Data Collection}\label{sec:app-teleoperation-data-collection}

\begin{figure}[t]
    \centering
    \includegraphics[width=\linewidth]{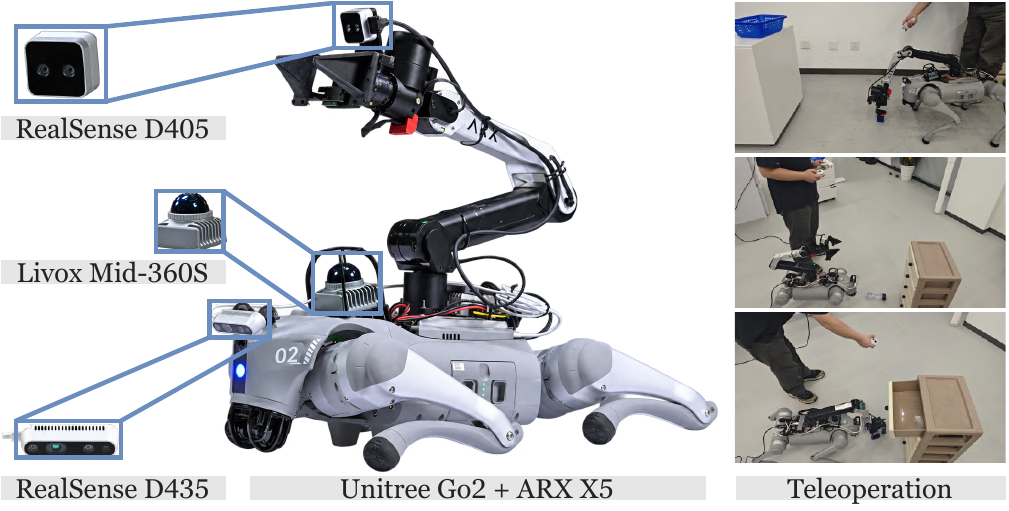}
    \caption{Hardware and teleoperation. Left: the Unitree Go2 with an ARX X5 arm, a wrist-mounted RealSense D405, a front-mounted RealSense D435, and a Livox Mid-360S LiDAR. Right: an operator collects demonstrations with handheld PICO VR controllers while staying near the robot.}
    \label{fig:teleoperation-hardware}
\end{figure}

\headindent \textbf{Robot and sensors.} Our platform combines a Unitree Go2 quadruped with a 6-DoF ARX X5 arm and a gripper (\Cref{fig:teleoperation-hardware}, left). A front-mounted RealSense D435 and a wrist-mounted RealSense D405 provide the two RGB views that are recorded in the demonstrations and given to the VLA. A Livox Mid-360S LiDAR supplies measurements for LiDAR--inertial odometry~\cite{fastlio2}. An onboard NVIDIA Orin runs the whole-body controller and the ROS2 control stack. The controller runs at $50\,\mathrm{Hz}$, while the arm and leg SDK control loops run at $500\,\mathrm{Hz}$.

\textbf{VR command interface.} The operator teleoperates the robot with PICO VR controllers (\Cref{fig:teleoperation-hardware}, right). We read controller poses and inputs through XRoboToolkit and publish commands over ROS2 at $30\,\mathrm{Hz}$. The left joystick controls the planar base velocity, the horizontal axis of the right joystick controls the yaw rate, and the right trigger sets the gripper opening from $0$ to $0.08\,\mathrm{m}$. The right grip button acts as a clutch. Pressing it records reference poses for the hand controller and the end effector. While the grip is held, the controller's displacement and rotation relative to its reference pose, converted from PICO to ROS coordinates, are applied to the end-effector reference pose to form the target, without scaling the translation. Releasing the grip freezes the end-effector target, so the operator can reposition the hand before pressing the grip again. The target frame follows the base translation in the horizontal plane and the base yaw, and its height is fixed when the clutch is engaged. $\pi_{\mathrm{wbc}}$ executes the pose targets and base-velocity commands.

\textbf{Demonstration recording.} We record the robot state at $50\,\mathrm{Hz}$, including all 18 joint positions, the gripper position, the arm-mount orientation, the realized and commanded end-effector poses in the world frame, and the base-velocity commands. The two cameras record RGB images independently at $30\,\mathrm{Hz}$, at $640\times480$ resolution for the wrist view and $424\times240$ for the front view. State samples and images are timestamped with the ROS clock and saved per episode, so the streams can be aligned afterward (\Cref{app:vla-data-processing}).

\textbf{Per-task variation.} We vary the initial conditions of each task across demonstrations so that the policy does not overfit to one scene. For mobile pick-and-place, we use different bottles and bottle-basket layouts, and the robot starts at different distances from the bottle. For mobile box pushing, the robot starts at different distances from the box, and pushes to the left and to the right are equally frequent. For drawer manipulation, we randomize the bottle position. Because grasping the handle requires fine alignment, we also record additional segments that start with the gripper near the handle and end with the grasp, so that the policy sees more examples of the final alignment. For mobile object transport, we place the cube at different positions and orientations in front of the robot. For clean whiteboard, we write words of varying length horizontally and vertically.

\section{VLA Training Details}\label{sec:app-vla-training-details}
\headindent This section describes how raw teleoperation episodes become VLA training samples (\Cref{app:vla-data-processing}) and how we post-train the two VLA backbones on them (\Cref{app:post-training}).

\subsection{Data Processing}
\label{app:vla-data-processing}

\subsubsection{Episode Formatting}
\headindent We convert each raw episode, which contains robot states and synchronized front- and wrist-camera streams, into training samples in three steps. First, we discard invalid control cycles, incomplete episodes, frames with missing camera observations, and episodes whose length falls outside $0.5$--$1.5\times$ the median of their collection (\Cref{tab:wbc-cleaning-thresholds}). Second, the wrist-camera timeline defines the samples: for each wrist frame, we pair the temporally nearest robot state (recorded at $50\,\mathrm{Hz}$) and front-camera frame. Third, we assemble each sample from the two images, the current proprioceptive state, the language instruction, and an action chunk of horizon $H=50$ ($1\,\mathrm{s}$) for $\pi_{0.5}$ or $H=40$ ($0.8\,\mathrm{s}$) for GR00T N1.7, omitting samples without a complete future horizon. We express pose targets relative to the chunk start rather than as adjacent-step increments, which gave a lower training loss. \Cref{tab:vla-layout} lists the resulting action and state layouts; the paragraphs below explain the design choices behind them.

\begin{table}[ht]
    \centering
    \caption{Layout of the two action interfaces and the proprioceptive state. Numbers in parentheses are dimensions.}
    \label{tab:vla-layout}
    \small
    \setlength{\tabcolsep}{3pt}
    \begin{tabularx}{\linewidth}{@{}>{\raggedright\arraybackslash}p{0.27\linewidth}c>{\raggedright\arraybackslash}X@{}}
        \toprule
        Quantity & Dim. & Layout \\
        \midrule
        End-effector action & 10 & translation (3), rotation vector (3), gripper (1), base command (3) \\
        Structured action & 12 & arm joints (6), arm-mount roll and pitch (2), gripper (1), base command (3) \\
        Proprioceptive state & 13 & arm-mount roll and pitch (2), base yaw (1), arm joints (6), gripper (1), previous base command (3) \\
        \bottomrule
    \end{tabularx}
\end{table}

\textbf{End-effector action interface.} When the VLA drives $\pi_{\mathrm{wbc}}$, it is supervised with the VR-commanded end-effector pose rather than the realized pose. The operator watched the robot and adjusted the command to compensate for its residual tracking error, so the commanded pose already contains this visual correction (\Cref{sec:teleop}). Let $\bm{T}^{W}_{E,0}$ be the world-frame end-effector pose at the start of a chunk, obtained from LiDAR odometry, and $\widehat{\bm{T}}^{W}_{E,k}$ the world-frame commanded pose at future step $k$. We express every target in the frame anchored at $\bm{T}^{W}_{E,0}$:
\begin{equation}
    \bm{T}^{E_0}_{E,k}
    =
    \left(\bm{T}^{W}_{E,0}\right)^{-1}
    \widehat{\bm{T}}^{W}_{E,k},
    \quad
    k \in \{0,\ldots,H-1\}.
    \label{eq:chunk-local-pose}
\end{equation}
This removes global position and heading while preserving the commanded motion relative to the chunk start. Rotations are stored as axis-angle vectors.

\textbf{Structured action interface.} When the VLA drives $\pi_{\mathrm{dwbc}}$, the arm joints, arm-mount roll and pitch, and gripper position are the \emph{realized} values recorded during collection, whereas the base command is the commanded velocity. Since $\pi_{\mathrm{dwbc}}$ tracks these targets precisely, replaying the realized motion reproduces what the operator achieved.

\textbf{Predicting base velocities.} Our controllers drive the Go2 through Unitree's low-level control service, which disables the high-level sport service and, with it, the measured base velocity. Rather than recovering the velocity by integrating IMU readings, we treat base-velocity prediction as sequence forecasting: given the previous velocity command and the current observations, the VLA predicts the next velocity commands. This way, training converges and the policy reproduces the operator's stopping behavior.

\textbf{Proprioceptive observations.} The state combines the arm-mount orientation (with yaw relative to the episode start), the measured arm joints, the gripper position, and the previous base command, which serves as the velocity surrogate described above. Leg joints are omitted because gait is handled by $\pi_{\mathrm{dwbc}}$.

\subsubsection{Quality Inspection and Data Cleaning}
\headindent After formatting, we remove samples with synchronization errors, static action windows, or action spikes, using the criteria in \Cref{tab:wbc-cleaning-thresholds}, where all norms are Euclidean. A sample is static only if the translation, rotation, gripper motion, and base command all stay below their thresholds throughout the first half of its action chunk. Spike detection examines the first action of each chunk and its change from the preceding sample. We remove every sample flagged by any check, which amounts to about 10\% of the samples in every task, mostly static windows.
\begin{table}[t]
    \centering
    \caption{Quality criteria for episode and sample filtering. A sample is flagged when a synchronization or spike threshold is exceeded, or when all static conditions hold over the window.}
    \label{tab:wbc-cleaning-thresholds}
    \small
    \setlength{\tabcolsep}{3pt}
    \begin{tabularx}{\linewidth}{@{}>{\raggedright\arraybackslash}p{0.27\linewidth}>{\raggedright\arraybackslash}p{0.29\linewidth}>{\raggedright\arraybackslash}X@{}}
        \toprule
        Check & Threshold & Flagging rule \\
        \midrule
        Episode length
            & $[0.5,1.5]\times$ median
            & Outside the collection-relative interval \\

        Image--state sync.
            & $0.05\,\mathrm{s}$
            & Timestamp residual exceeds threshold \\

        Front--wrist sync.
            & $0.10\,\mathrm{s}$
            & Timestamp residual exceeds threshold \\

        Static window
            & $0.5H$ steps
            & All four static conditions hold in the leading window \\

        EE translation
            & $2\,\mathrm{mm}$
            & $\|\Delta\bm{p}_k\|_2$ remains below threshold \\

        EE rotation
            & $0.01\,\mathrm{rad}$
            & $\|\Delta\bm{\phi}_k\|_2$ remains below threshold \\

        Gripper motion
            & $2\,\mathrm{mm}$
            & Consecutive gripper changes remain below threshold \\

        Base motion
            & $0.02$
            & Base-command norm remains below threshold \\

        Translation spike
            & $0.20\,\mathrm{m}$
            & First-step translation norm exceeds threshold \\

        Rotation spike
            & $0.80\,\mathrm{rad}$
            & First-step rotation norm exceeds threshold \\

        Gripper anomaly
            & $[-0.005,0.10]\,\mathrm{m}$
            & Value is outside the valid interval \\

        Gripper jump
            & $0.04\,\mathrm{m}$
            & Change from the preceding sample exceeds threshold \\

        Base-command spike
            & $2.0$
            & First-step command norm exceeds threshold \\

        Base-command jump
            & $1.0$
            & Change from the preceding sample exceeds threshold \\
        \bottomrule
    \end{tabularx}
\end{table}

\subsection{VLA Post-Training}
\label{app:post-training}

\headindent We post-train $\pi_{0.5}$~\cite{pi05} and GR00T N1.7~\cite{gr00tN1} with their native training stacks, adding data loaders for the precomputed action chunks of \Cref{app:vla-data-processing} and optional training-time real-time chunking (TTRTC)~\cite{ttrtc}. Both models receive the two RGB images, the language instruction, the proprioceptive state, and an action chunk in either the end-effector or the structured interface.

\textbf{Training-time real-time chunking.} TTRTC implements the prefix masking of \Cref{eq:flow-matching}. For each sample, we draw a prefix length $d$, give the first $d$ actions of the chunk without noise, and exclude them from the loss, while the remaining actions follow the backbone's own noising procedure. The prefix stands for the actions executed while the next chunk is computed. One inference with 5 flow-integration steps for $\pi_{0.5}$ takes $100\,\mathrm{ms}$ on an RTX 4090, or 5 control steps of $20\,\mathrm{ms}$, so we sample $d$ uniformly from $\{0,\dots,10\}$ to cover up to twice this delay.

\textbf{Training data.} Single-task policies use 400 demonstrations for mobile pick-and-place, mobile box pushing and mobile object transport, and 500 for drawer manipulation and clean whiteboard. The multi-task policy of \Cref{sec:exp-capability} is trained on all five datasets, and the mid-training model on mobile pick-and-place and drawer manipulation. When a run mixes tasks, task $i$ is sampled with probability proportional to $n_i^{0.43}$, where $n_i$ is its number of frames. This sublinear weighting keeps the largest dataset from dominating while still favoring tasks with more data.

\textbf{Optimization.} \Cref{tab:post-training-hyperparameters} lists the settings of each run. All runs use a cosine learning-rate schedule with 300 warm-up steps that decays to $0.1\times$ the peak rate. Multi-task and mid-training runs decay over their full length, whereas both single-task settings share a 12{,}000-step decay horizon, so the 400-demonstration runs stop before the schedule ends.

\begin{table}[t]
    \centering
    \caption{Post-training settings. Demos is the number of demonstrations in one training run.}
    \label{tab:post-training-hyperparameters}
    \small
    \setlength{\tabcolsep}{4pt}
    \begin{tabularx}{\linewidth}{@{}Xcccc@{}}
        \toprule
        Run & Demos & Steps & Batch & Peak LR \\
        \midrule
        Single-task & $400$ & $9{,}000$ & $64$ & $2.5\times10^{-5}$ \\
        Single-task & $500$ & $12{,}000$ & $64$ & $2.5\times10^{-5}$ \\
        Multi-task & $2{,}200$ & $26{,}000$ & $128$ & $5.0\times10^{-5}$ \\
        Mid-training & $900$ & $15{,}000$ & $128$ & $5.0\times10^{-5}$ \\
        \bottomrule
    \end{tabularx}
\end{table}

\textbf{Backbone-specific settings.} For $\pi_{0.5}$, we normalize states and actions with quantile normalization, which maps the 1st and 99th percentiles of each dimension to $-1$ and $1$, and fine-tune all parameters of the VLM backbone and the action expert. The official GR00T recipe instead uses min-max normalization, but with it GR00T produced large actions that destabilized the base, so we apply the same quantile normalization. For GR00T, we freeze the language and vision backbones and train the multimodal projector, the diffusion transformer, and the vision-language layer normalization (VLLN). Both models are trained in bfloat16.

\textbf{Deployment.} At deployment, both models receive the same camera mapping, instruction and proprioceptive state as in training. With TTRTC enabled, the client also sends the previously issued chunk and how many of its actions were executed during inference, and the model conditions on these executed actions as the noise-free prefix. We use 5 flow-integration steps for both $\pi_{0.5}$ and GR00T N1.7. Inference takes $100\,\mathrm{ms}$ for $\pi_{0.5}$ and $83\,\mathrm{ms}$ for GR00T N1.7 on an RTX 4090.

\end{document}